\documentclass{article}

\usepackage{microtype}
\usepackage{graphicx}
\usepackage{booktabs} 
\usepackage{amsmath}
\usepackage{float}
\usepackage{subcaption}
\usepackage{multirow}
\usepackage{amsfonts}
\DeclareMathOperator{\Tr}{Tr}
\usepackage{hyperref}

\usepackage[accepted]{icml2021}

\icmltitlerunning{How to Stay Curious while Avoiding Noisy TVs}

\begin{document}

\twocolumn[
\icmltitle{How to Stay Curious while avoiding Noisy TVs \protect\\ using Aleatoric Uncertainty Estimation}



\icmlsetsymbol{equal}{*}

\begin{icmlauthorlist}
\icmlauthor{Augustine N.~Mavor-Parker}{uclai}
\icmlauthor{Kimberly A.~Young}{uclcdb,bu}
\icmlauthor{Caswell Barry}{uclcdb,equal}
\icmlauthor{Lewis D.~Griffin}{uclcs,equal}
\end{icmlauthorlist}

\icmlaffiliation{uclcs}{Department of Computer Science, University College London, UK}
\icmlaffiliation{uclai}{Centre for Artificial Intelligence, University College London, UK}
\icmlaffiliation{uclcdb}{Department of Cell and Developmental Biology, University College London, UK}
\icmlaffiliation{bu}{Boston University, Center for Systems Neuroscience, Graduate Program for Neuroscience, USA}
\icmlcorrespondingauthor{Augustine Mavor-Parker}{a.mavor-parker@cs.ucl.ac.uk}

\icmlkeywords{Machine Learning, ICML}

\vskip 0.3in
]



\printAffiliationsAndNotice{\icmlEqualContribution} 

\begin{abstract}
When extrinsic rewards are sparse, artificial agents struggle to explore an environment. Curiosity, implemented as an intrinsic reward for prediction errors, can improve exploration but it is known to fail when faced with action-dependent noise sources (‘noisy TVs’).  In an attempt to make exploring agents robust to noisy TVs, we present a simple solution: aleatoric mapping agents (AMAs). AMAs are a novel form of curiosity that explicitly ascertain which state transitions of the environment are unpredictable, even if those dynamics are induced by the actions of the agent. This is achieved by generating separate forward predictions for the mean and aleatoric uncertainty of future states, with the aim of reducing intrinsic rewards for those transitions that are unpredictable. We demonstrate that in a range of environments AMAs are able to circumvent action-dependent stochastic traps that immobilise conventional curiosity driven agents. Furthermore, we demonstrate empirically that other common exploration approaches---previously thought to be immune to agent-induced randomness---can be trapped by stochastic dynamics. Code to reproduce our experiments is \href{https://github.com/self-supervisor/How_to_stay_curious_while_avoiding_noisy_TVs}{provided}.
\end{abstract}

\section{Introduction}\label{submission}

Efficient exploration is a central problem
in reinforcement
learning. Agents need to be capable of finding novel information
without depending on extrinsic rewards to shepherd them through the state space 
of a given environment
(e.g. \citet{sutton2018reinforcement,
    pathak2017curiosity,
burda2018exploration}, see
\citet{weng2020exploration} for a
review). A notable exploration method that
effectively deals
with
sparse rewards is curiosity driven
learning---where
agents are equipped with a self-supervised
forward prediction
model
that employs prediction errors as
intrinsic rewards
\citep{schmidhuber1991possibility, pathak2017curiosity,
schmidhuber1991adaptive}.
Curiosity
is
built upon the intuition that in
unexplored regions of the
environment, the forward prediction error
of the agent's
internal
model will be large
\citep{schmidhuber1991possibility, pathak2017curiosity}. As a result,
agents are rewarded for visiting regions
of the state space
that
they have not previously occupied.
If, however, a particular
state
transition is impossible to predict, it
will trap a curious
agent
\citep{burda2018exploration,
schmidhuber1991adaptive}. This is
referred to as the noisy TV problem (e.g.
\citet{burda2018exploration, schmidhuber1991adaptive}), the
etymology being that a
naively
curious agent could dwell on the
unpredictability of a noisy
TV
screen. 

Several existing curiosity-like
methods
\cite{burda2018exploration,
    pathak2017curiosity,
pathak19disagreement} aim to
avoid noisy TVs or
``stochastic traps''
\cite{shyam2019model}. Nevertheless, employing
dynamics based prediction errors as
intrinsic rewards is difficult as
current methods either 
fail when stochastic traps
are
action-dependent,
or
require an ensemble of
dynamics models
\cite{pathak2017curiosity,
    pathak19disagreement,
    shyam2019model,
burda2018large}. Even if ensemble methods are available, we demonstrate that they cannot reliably overcome the allure of observing random observations. Additionally, we find that random network distillation---a dynamics-free exploration technique usually assumed to be robust to stochasticity---is also susceptible to noisy TVs. Fundamentally, popular intrinsic reward approaches are vulnerable to the never ending novelty of a noisy TV.

We present a simple solution to the noisy TV problem---instead of only predicting the next state, we also predict its variance (i.e it's \textit{aleatoric uncertainty} \cite{kendall2017uncertainties}). The uncertainty 
of a statistical model can be 
described as the sum of two
\textit{theoretically} distinct
types
of uncertainty: epistemic uncertainty and
aleatoric
uncertainty
(e.g. \citet{hora1996aleatory}, see \citet{hullermeier2019aleatoric} for a review).
Epistemic uncertainty
measures
the errors of a model's prediction
that can be minimised 
with additional experience and learning
\cite{hullermeier2019aleatoric}.
As
a result, an agent using epistemic uncertainties as
intrinsic rewards tends to value dynamics 
it has not previously encountered, 
and hence cannot predict accurately, but could 
learn to predict in the future
(e.g. \citet{osband2016deep}). More concretely, epistemic uncertainty 
can be considered to be the ``expected information gain'' of observing the next predicted 
state \cite{mukhoti2021deterministic}. On the
other hand, prediction
errors
that are due to aleatoric uncertainties
are, by definition,
a result of
unpredictable
processes
\cite{hullermeier2019aleatoric}.
Therefore, any agent that receives 
intrinsic rewards for aleatoric dynamics 
risks being trapped, as
exemplified by the noisy TV problem
\cite{schmidhuber1991adaptive, burda2018large}. By predicting aleatoric uncertainties, our curious agents are able to disregard stochastic dynamics if they are consistent with the agent's predicted variance---avoiding the trap of noisy TVs. Our contributions are summarised as follows: 

\begin{enumerate}
    \item We benchmark the performance of existing exploration techniques, highlighting that they are more vulnerable to stochasticity than previously assumed
    \item We present a novel form of curiosity that can operate proficiently in  exploration benchmark environments in the presence of a noisy TV, while still preserving exploration performance without a noisy TV
    \item We show that even in the famously deterministic domain of Atari, natural sources of randomness exists
\end{enumerate}

Finally, we also highlight the connections of AMAs to experimental neuroscience. Our approach to resisting stochasticity is both inspired by and builds upon proposals developed within neuroscience \cite{angela2005uncertainty}, that suggest expected uncertainties in predictions of future states are signalled by the modulation of cortical acetylcholine in the mammalian brain. The implications for neuroscience and potential animal experiments are included in the discussion (section \ref{ACh}) and Appendix \ref{bandit}.

\section{Background}
\subsection{Epistemic and Aleatoric
Uncertainties}

Estimating the epistemic uncertainty surrounding 
future states would be an ideal basis for 
a curious agent but tractable epistemic uncertainty 
estimation with high dimensional data is 
an unsolved problem (see \citet{gal2016uncertainty} for an introduction to the field).
We implicitly incentivise agents to seek
epistemic
uncertainties
by 
removing the aleatoric component from the
total prediction error. More specifically, fundamental to our model is the maximum likelihood approach from \citet{kendall2017uncertainties}, which we use to predict aleatoric uncertainties based on the input. We then subtract these aleatoric uncertainties from prediction errors, which is a novel approach for efficiently estimating epistemic uncertainties.

There are
similar methods that separate epistemic and
aleatoric
uncertainties in return predictions
\citep{clements2019estimating}, or within 
a latent variable model \citep{depeweg2018decomposition}---allowing for the construction of policies
that are rewarded
for
exploring their environments and punished
for experiencing
aleatoric uncertainty. However, as far
as we are aware, we
are the first to compute 
aleatoric uncertainties
within a {\it scalable} curiosity
framework 
to remove {\it intrinsic rewards} for those
state transitions with aleatoric uncertainty, which implicitly rewards agents for experiencing epistemic 
uncertainties. However, we do note that a similar approach was announced shortly after ours \citep{jain2021deup},
which shows that epistemic uncertainties can be estimated by subtracting aleatoric uncertainty from a predicted prediction error. Our approach is simpler, rather than trying to predict prediction errors, we use the implicitly calculated epistemic uncertainty as 
intrinsic rewards online.
\subsection{Curiosity and Intrinsic
    Motivation in
    Reinforcement
Learning}
Curiosity-driven
\citep{pathak2017curiosity} agents 
assign value to states of the environment
that they 
deem to be ``interesting'' \citep{still2012information, Schmidhuber97whatsinteresting?}. 
How a curiosity based method computes
whether a state is 
``interesting''
\citep{still2012information, Schmidhuber97whatsinteresting?} is usually
its
defining 
characteristic. The original formulation of 
curiosity used prediction errors directly 
as intrinsic rewards \citep{schmidhuber1991possibility}. The noisy TV problem 
quickly emerged when using this na\"ive approach in stochastic environments \citep{schmidhuber1991adaptive}.
In order to evade the allure of stochastic traps, the first proposed solution to the noisy TV problem 
implements ``interesting'' \citep{still2012information, Schmidhuber97whatsinteresting?} 
as prediction errors that reduce 
over time \citep{schmidhuber1991adaptive, kaplan2007search}. Others consider ``interesting'' \citep{still2012information, 
Schmidhuber97whatsinteresting?} to mean a high dependency between 
present and future states and actions (i.e. ``interesting'' things are
predictable \citep{still2012information} or 
controllable \citep{mohamed2015variational}).

Inverse dynamics feature (IDF) curiosity
\citep{pathak2017curiosity} 
rejuvenated interest in using one step prediction errors as intrinsic rewards. 
IDF curiosity avoids stochastic traps 
by computing prediction errors with features
that aim to only contain information
concerning stimuli the agent
can affect \citep{pathak2017curiosity}. 
Further experiments \citep{burda2018large} 
showed that simple one-step  
prediction errors also work 
effectively within a random representation
space generated 
by feeding state observations through a
randomly initialised 
network. \citet{burda2018large} also showed the
(IDF) approach is vulnerable to action-dependent noisy
TVs---demonstrated
by giving the agent a `remote control' to
a noisy TV in the
environment 
that could induce unpredictable
environment transitions. 
This motivated random
network distillation (RND)
\citep{burda2018exploration}, which 
removes 
dynamics from the prediction problem 
altogether---instructing a
network to learn to
predict
the 
output of another fixed randomly
initialised network at 
each state, using the resulting error as
intrinsic rewards. RND persists as
a key component in state of the art algorithms 
deployed in high dimensional state spaces \citep{badia2020never}. 

Other exploration methods explicitly
leverage 
uncertainty quantification for
exploration. 
The canonical approach is
``optimism under uncertainty'', which in
its most basic form 
means weighting the value of state-actions
pairs inversely to
the number of times they have been experienced
\citep{sutton2018reinforcement}(p. 36).
Known as count 
based methods \citep{strehl2008analysis,
bellemare2016unifying}, this approach was
shown to reliably evade
noise sources in minigrid environments
\cite{Raileanu2020RIDE:}.
However, 
it is not feasible to count state 
visitations in many environments where
there is a 
large number of unique states \cite{bellemare2016unifying}. 
``Pseudo-count'' methods exchange tabular 
look up tables for density models to
estimate an
analogous 
intrinsic reward to counts in large state
spaces
\cite{bellemare2016unifying}---related to density models, \cite{kim2019curiosity} use state ``compressibility'' as intrinsic rewards. 

Attempts have been made to reward epistemic uncertainty explicitly. This typically requires a posterior
distribution over model
parameters, which 
is intractable without approximations such as ensembles 
or variational inference (e.g. \citet{houthooft2016vime}). \citet{osband2016deep}
instantiated an ensemble
\citep{lakshminarayanan2016simple} 
into the final layer of a deep Q-network---rewarding its 
agents for epistemic value uncertainty.
\citet{pathak19disagreement} use the variance of ensemble 
predictions being used as intrinsic rewards, while \citet{shyam2019model} 
reward experience of
epistemic uncertainty within 
an ensemble of environment models. Lastly, uncertainty estimation methods have recently been developed that enforce a smoothness constraint in the representation space \citep{mukhoti2021deterministic, van2021feature}---allowing for sensible estimations of uncertainty to be made from learned representations---but these approaches have not yet been adopted in reinforcement learning.
\section{Method}
Our method operates in an environment defined as a Markov
decision process that consist of states $s
\in \mathcal{S}$,
actions $a \in \mathcal{A}$, and rewards
$r \in \mathcal{R}
\subset
\mathbb{R} $
\citep{sutton2018reinforcement}.
At each
timestep
$t$
the agent selects an action via a
stochastic policy
$a_{t} \sim \pi (\cdot | s_{t})$
\citep{szepesvari2010algorithms} and then
receives
a
reward $r_{t+1}$
and state $s_{t+1}$ generated via
the transition
function
$p(\mathbf{s}_{t+1},
r_{t+1}|\mathbf{s}_{t}, a_{t})$ of
the environment
\citep{sutton2018reinforcement}. The
objective
of
the agent is to learn a
stochastic
policy $\pi$, parametrised by $\xi$,
which aims to maximise the
expectation of the sum of discounted
future rewards (e.g.
\citet{mnih2016asynchronous}).
\begin{equation}
\underset{\pi_{\xi}}{\mbox{max}}\;
\mathbb{E}_{\pi_{\xi}}
\left[
\sum_{k=0}^{T} \gamma^{k}r_{t+k} \right]
\end{equation}
Where $T$ is the episode length and
$\gamma$ is the discount
factor. Following other curiosity based
methods, the total
reward
is the sum of the intrinsic
reward provided by the intrinsic reward
module of the agent
and
the extrinsic reward provided by the
environment (e.g.
\citet{pathak2017curiosity,
    badia2020never, Raileanu2020RIDE:,
burda2018exploration}).
\begin{equation}\label{intrinsic_reward_equation}
    r_{t} = \beta r_{t}^{i} + r_{t}^{e}
\end{equation}
Where the superscripts $i$ and $e$ indicate intrinsic and
extrinsic rewards,
and $\beta$ is a hyperparameter that regulates the influence of
intrinsic
rewards on the policy. In previous works \citep{burda2018large}, the intrinsic reward 
$r_{t}^{i}$ is equal to the mean
squared forward prediction error of a curiosity module. To avoid stochastic traps
we subtract the aleatoric uncertainty---which is constrained to have a diagonal covariance \citep{kendall2017uncertainties}---from the prediction error, 
so that agents are 
not surprised by transitions that were previously learnt to be unpredictable.
\begin{equation}\label{eqn:intrinsic_reward_function}
r^{i}_{t} =\|\mathbf{s}_{t+1}-
{\mathbf{\hat{\mu}}}_{t+1}\|^{2} - \eta \Tr(\hat{\mathbf{\Sigma}}_{t+1})
\end{equation}
\noindent Where ${\hat{\mu}}_{t+1}$ is the predicted mean of the next state, 
$\hat{\mathbf{\Sigma}}_{t+1}$ is the predicted aleatoric uncertainty of the next state and
$\eta$ is a hyperparameter that regulates by how much the 
predicted uncertainty of the next state effects intrinsic rewards. 
To learn to predict the mean of the
next state $\hat{\mu}_{t+1}$ and its aleatoric uncertainty $\hat{\mathbf{\Sigma}}_{t+1}$, 
we follow \citet{kendall2017uncertainties}---fitting a diagonal 
covariance Gaussian distribution to the 
elements of the next state. The predictions are made by a double-headed neural 
network---with a mean prediction head 
$\mathbf{f}$ parameterised by $\theta$ and a variance prediction head $\mathbf{g}$ 
parametrised by $\phi$.
As employed in previous works, the separate heads of the double-headed deep 
network share feature 
extracting parameters \citep{kendall2017uncertainties}.
The prediction network performs \textit{heteroscedastic} aleatoric uncertainty estimation 
\citep{kendall2017uncertainties}, 
which in a reinforcement learning context means the prediction heads are conditioned on the current state and action,
\begin{equation}\label{eqn:proportionality_expression}
    p(\mathbf{s}_{1:N} | \theta, \phi) =
  \prod_{t=1}^{N}\mathcal{N}(\mathbf{s}_{t+1};
  \mathbf{f}_{\theta}(\mathbf{s}_{t}, \mathbf{a}_{t}),
  \mathbf{g}_{\phi}(\mathbf{s}_{t}, \mathbf{a}_{t}))
\end{equation}
\begin{figure*}[tp]
    \centering
    \subfloat[\centering]{{\includegraphics[width=8cm]{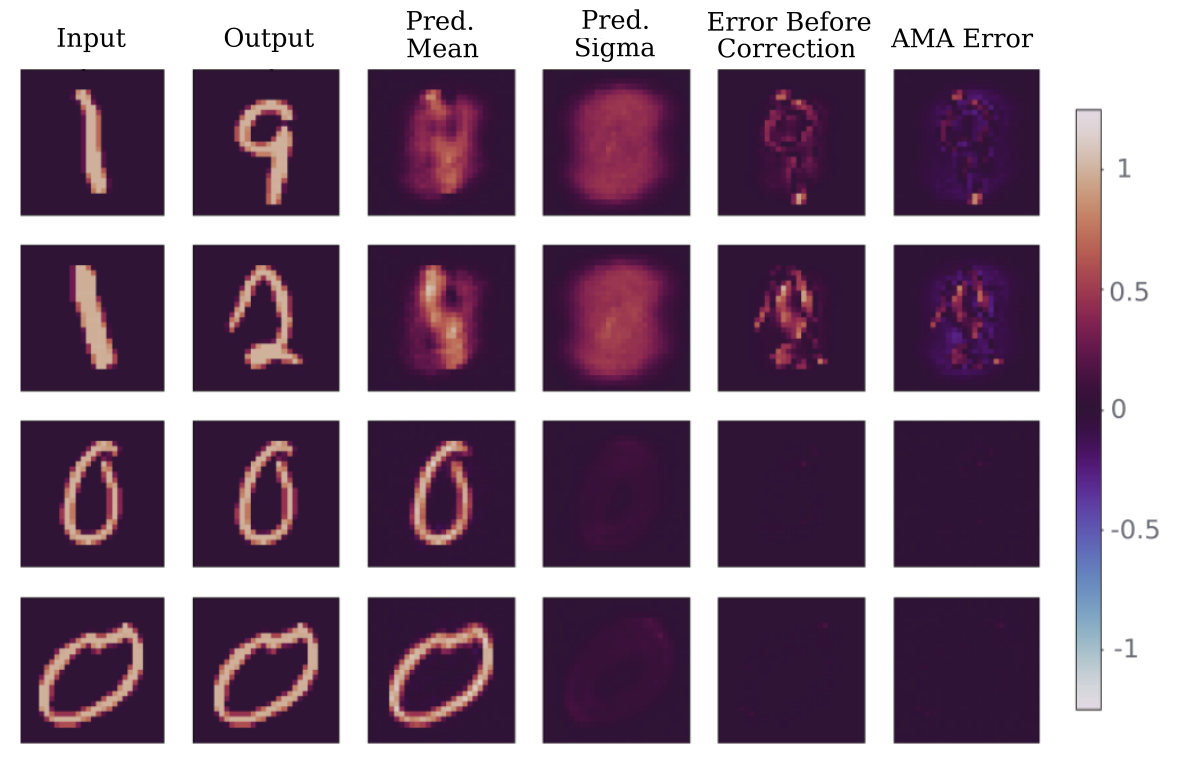} }}
    \subfloat[\centering]{{\includegraphics[width=8cm]{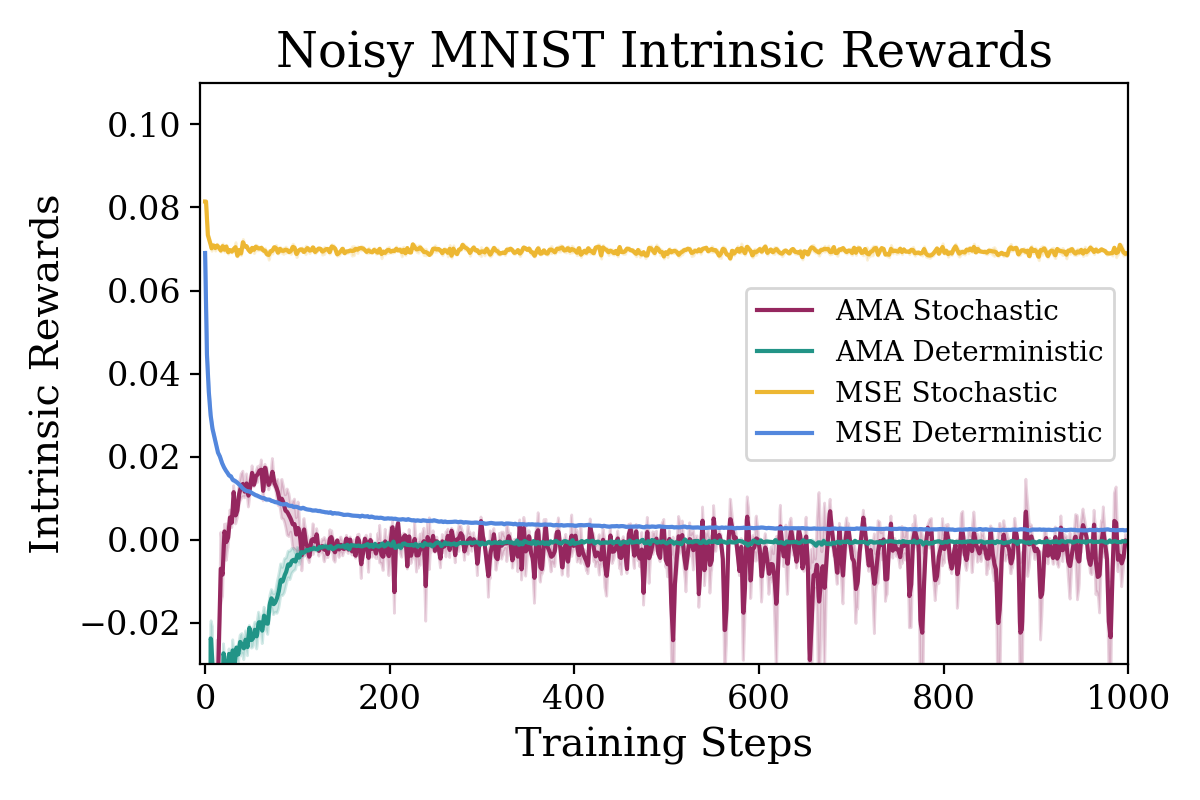}}}
    \caption{AMAs can learn to ignore stochastic transitions. 
            (a) Example transitions from the
            Noisy MNIST environment
            along with associated predictions. The top two
            rows show stochastic transitions
            where AMA's predicted variance is high in the
            majority of the image
            allowing intrinsic reward to be small despite the stochastic 
            transition. 
            (b) Two reward curves for MSE and AMA
            are plotted where stochastic is the
            $1\rightarrow\{2,...,9\}$
            transitions and deterministic
            is the $0\rightarrow0$ transitions.}
\end{figure*}
\noindent where $N$ is the total number of states observed during training. 
While \citet{kendall2017uncertainties} use {\it{maximum a posteriori}}
inference with a
zero-mean
Gaussian prior on the network parameters
$\{\theta, \phi\}$, we found empirically that the resulting regularisation
terms in the cost function did not improve results. Accordingly, we 
simply perform maximum likelihood estimation with the likelihood 
presented in Equation (\ref{eqn:proportionality_expression}),
resulting in the following cost function \citep{kendall2017uncertainties}.
\begin{equation}\label{eqn:heteroscedastic_objective}
\begin{aligned}
    \mathcal{L}_{t+1}(\theta,
    \phi)={}(\mathbf{s}_{t+1}-
        \hat{\mathbf{\mu}}_{t+1})^\top \hat{\mathbf{\Sigma}}_{t+1}^{-1} (\mathbf{s}_{t+1}-
        \hat{\mathbf{\mu}}_{t+1}) + \\ \lambda\log
        (\det(\hat{\mathbf{\Sigma}}_{t+1}))
\end{aligned}
\end{equation}
The first term is the familiar mean squared error \textit{divided by
the uncertainty} $\hat{\mathbf{\Sigma}}_{t+1}$. The second
term blocks the explosion of
predicted aleatoric
uncertainties
\citep{kendall2017uncertainties}.
We follow \citet{kendall2017uncertainties}'s prescription
of estimating $\log
\mathbf{\Sigma}$ instead of $\mathbf{\Sigma}$ to ensure stable optimisation. Furthermore, the
hyperparameter
$\lambda$ was added to adjust the model's
aleatoric uncertainty budget (e.g.
\citet{depeweg2018decomposition,
clements2019estimating,
eriksson2019epistemic}). We use the predicted mean and 
aleatoric uncertianty of the next state---which are being 
learned online with Equation (\ref{eqn:heteroscedastic_objective})---to compute intrinsic 
rewards according to Equation (\ref{eqn:intrinsic_reward_function}).
Lastly, we would like to highlight that the policy network is separate to the state prediction 
network as in other curiosity based methods \citep{pathak2017curiosity}.

\section{Experiments}

The purpose of this work is to improve the exploration capabilities of deep reinforcement learning algorithms in stochastic environments where current methods can fail catastrophically. As we are interested in exploration, we measure exploration directly by calculating an agent's environment coverage (when possible). Details of the noisy TVs used---including TVs we add to environments as well as a natural noisy TV in Atari---are contained within each subsection. Extra details such as the hyperparameters and architectures used are in Appendix \ref{implementation_details}. Shaded regions are standard error of the mean and we use 5 seeds for each method, except minigrid where we use 10 seeds and the MNIST experiments where only 3 were necessary.
\subsection{Noisy MNIST}
First we completed a supervised learning task, similar
to the noisy MNIST environment introduced
by \citet{pathak19disagreement}. The
environment does not
elicit any actions from an agent. Instead,
the prediction network simply needs to learn one 
step mappings between pairs of MNIST handwritten digits. 
The first images in the pairs are randomly selected 
0s or 1s. When the first image is a 0 then the 
second image is the exact same image (these are the 
deterministic transitions).
When the first image is a 1, then the second 
image is a random digit from 2-9 (these are the stochastic 
transitions).
A prediction
model
capable of avoiding noisy TVs should
eventually learn to compute
equal
intrinsic rewards for both types of
transitions
\citep{pathak19disagreement}.

We trained two different neural networks on this task (adapted from \citet{autoencoder}), one
with a mean squared error (MSE) loss function---as a baseline---and the other with the AMA loss function
\citep{kendall2017uncertainties}. The networks are
equivalent except that the AMA network has two
prediction heads. Both networks contain skip 
connection from the input layer to the output layer and were 
optimised with Adam \citep{kingma2014adam} at a learning rate of
0.001 and a batch size of 32. The uncertainty budget hyperparameter $\lambda$
and the uncertainty weighting hyperparameter $\eta$ were set to 1 for the AMA network. 
\begin{figure*}[tp]\label{fig:minigrid}
\centering
\begin{minipage}{0.55\textwidth}
\begin{subfigure}{0.45\textwidth}
\centering
\includegraphics[width=\linewidth]{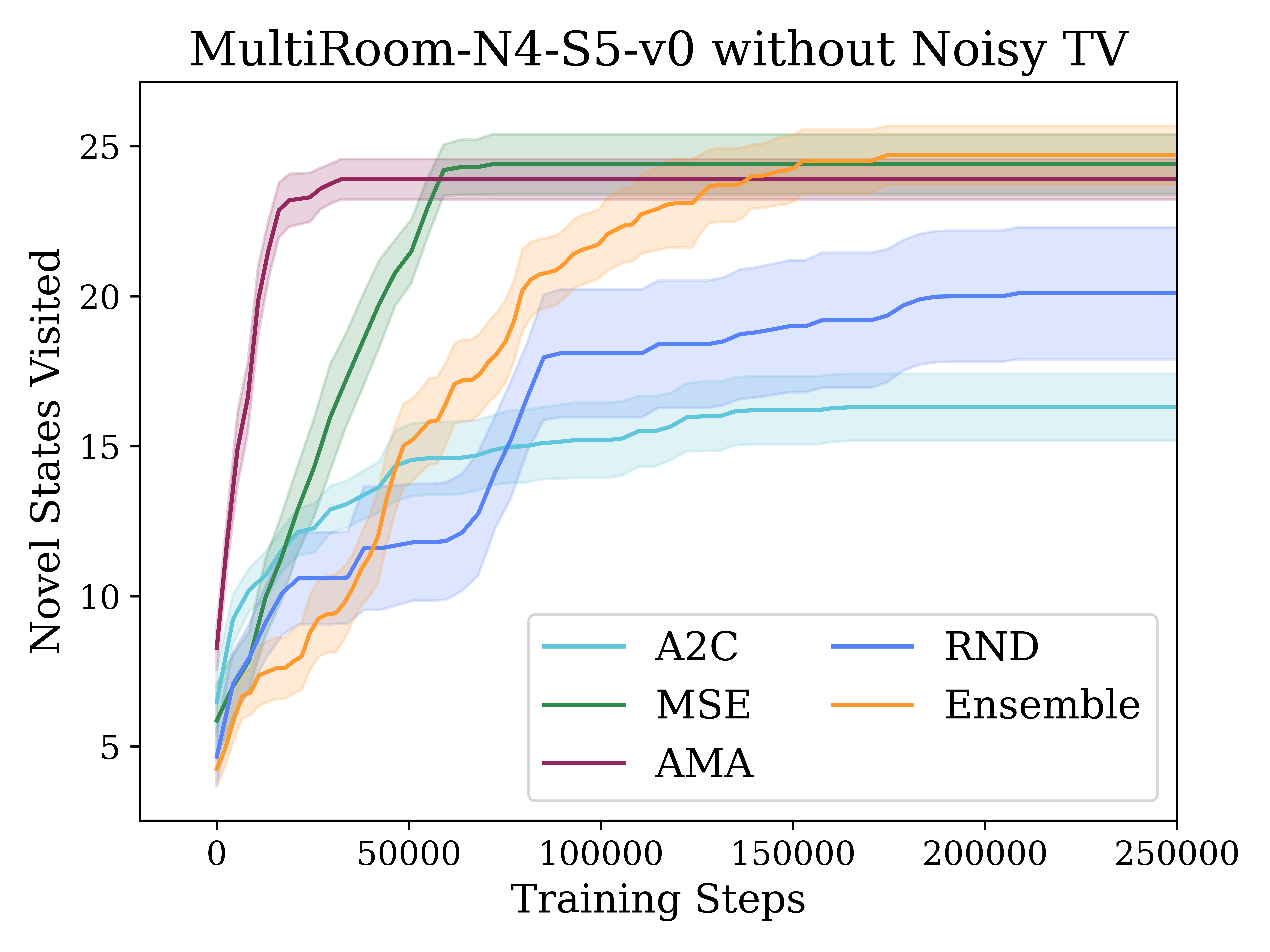}
\caption{} 
\end{subfigure}
\hfill
\begin{subfigure}{0.49\textwidth}
\centering
\includegraphics[width=\linewidth]{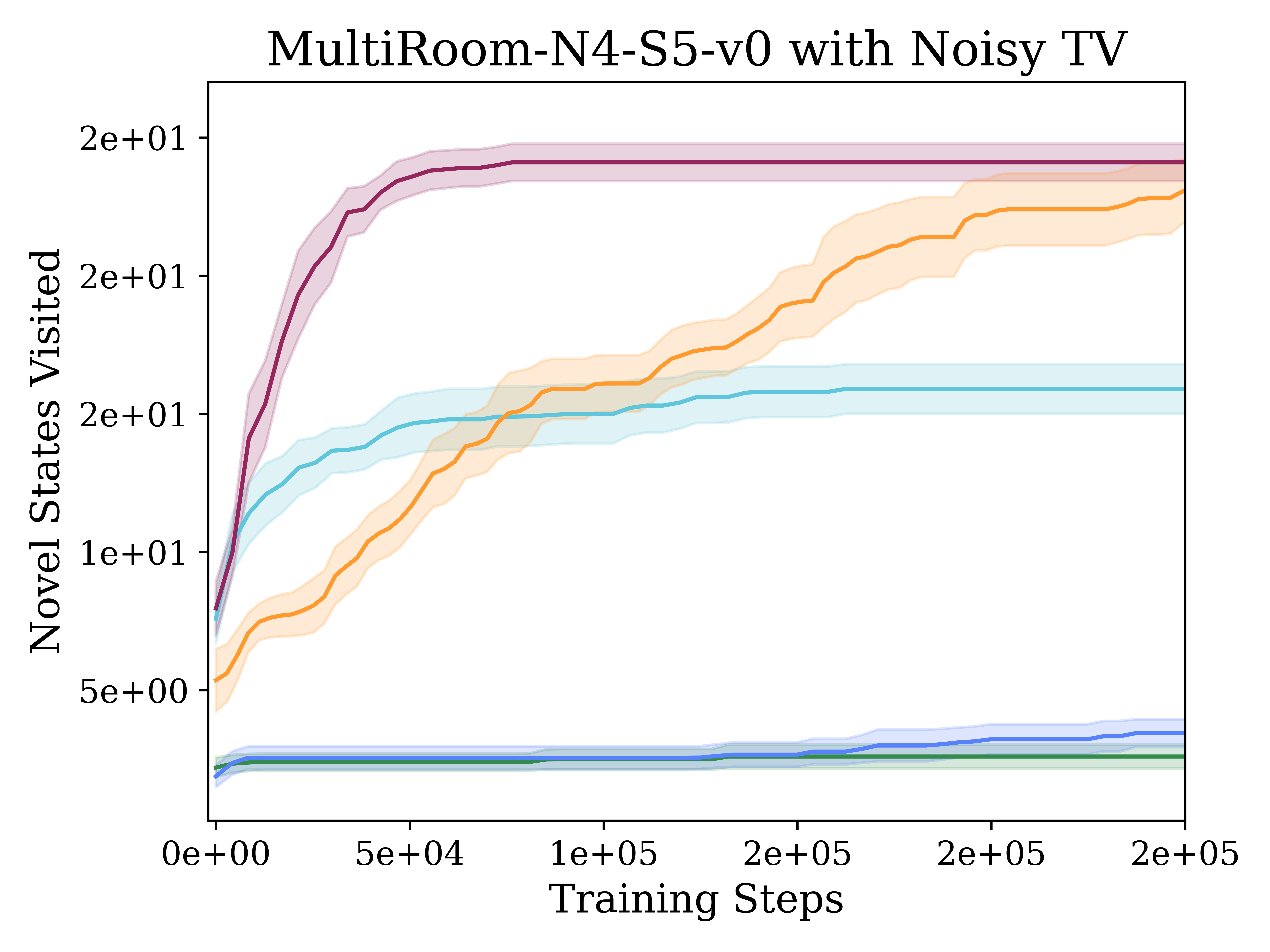}
\caption{} 
\end{subfigure}

\bigskip 
\begin{subfigure}{0.49\textwidth}
\centering
\includegraphics[width=\linewidth]{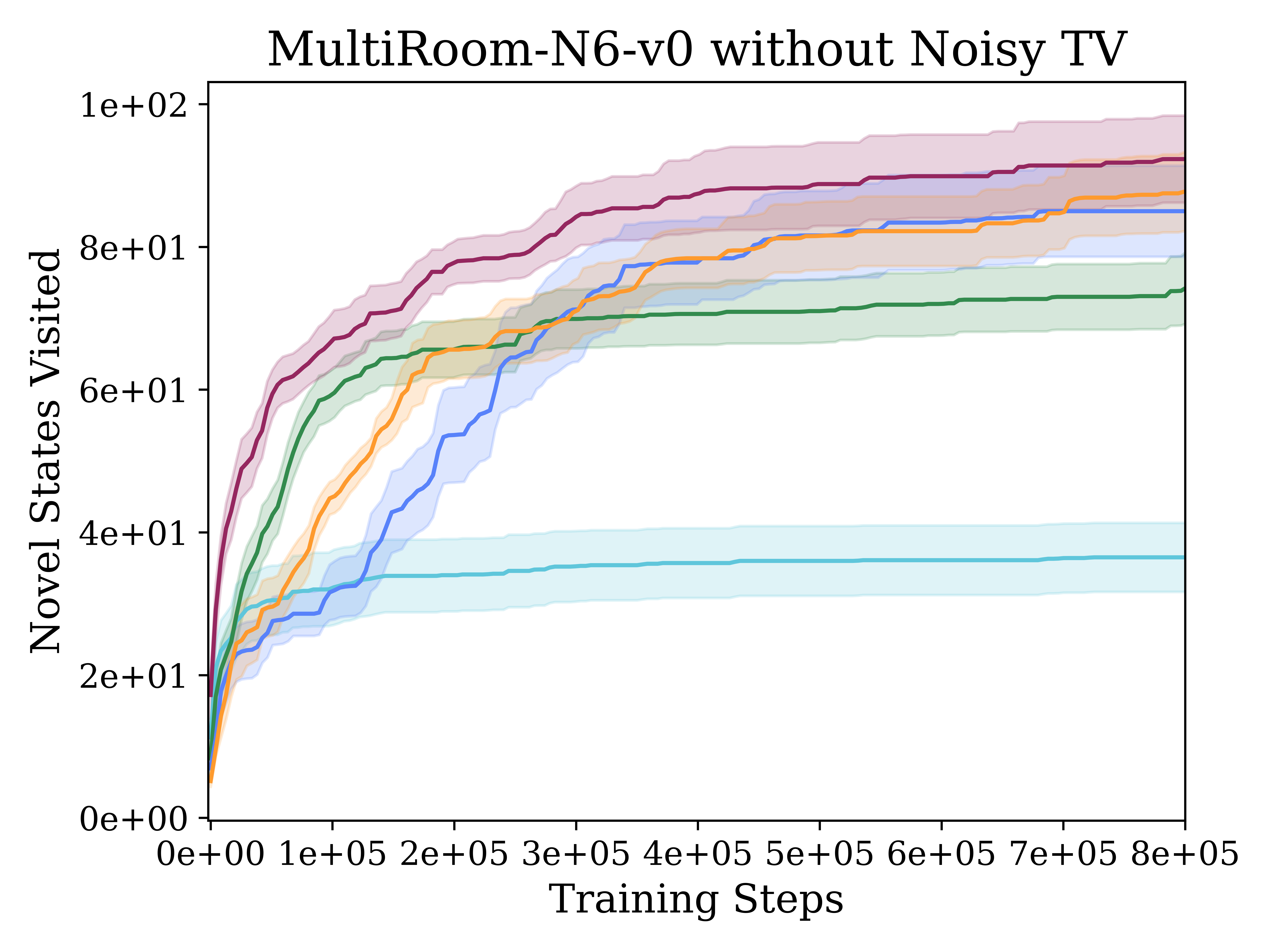}
\caption{} 
\end{subfigure}
\hfill
\begin{subfigure}{0.49\textwidth}
\centering
\includegraphics[width=\linewidth]{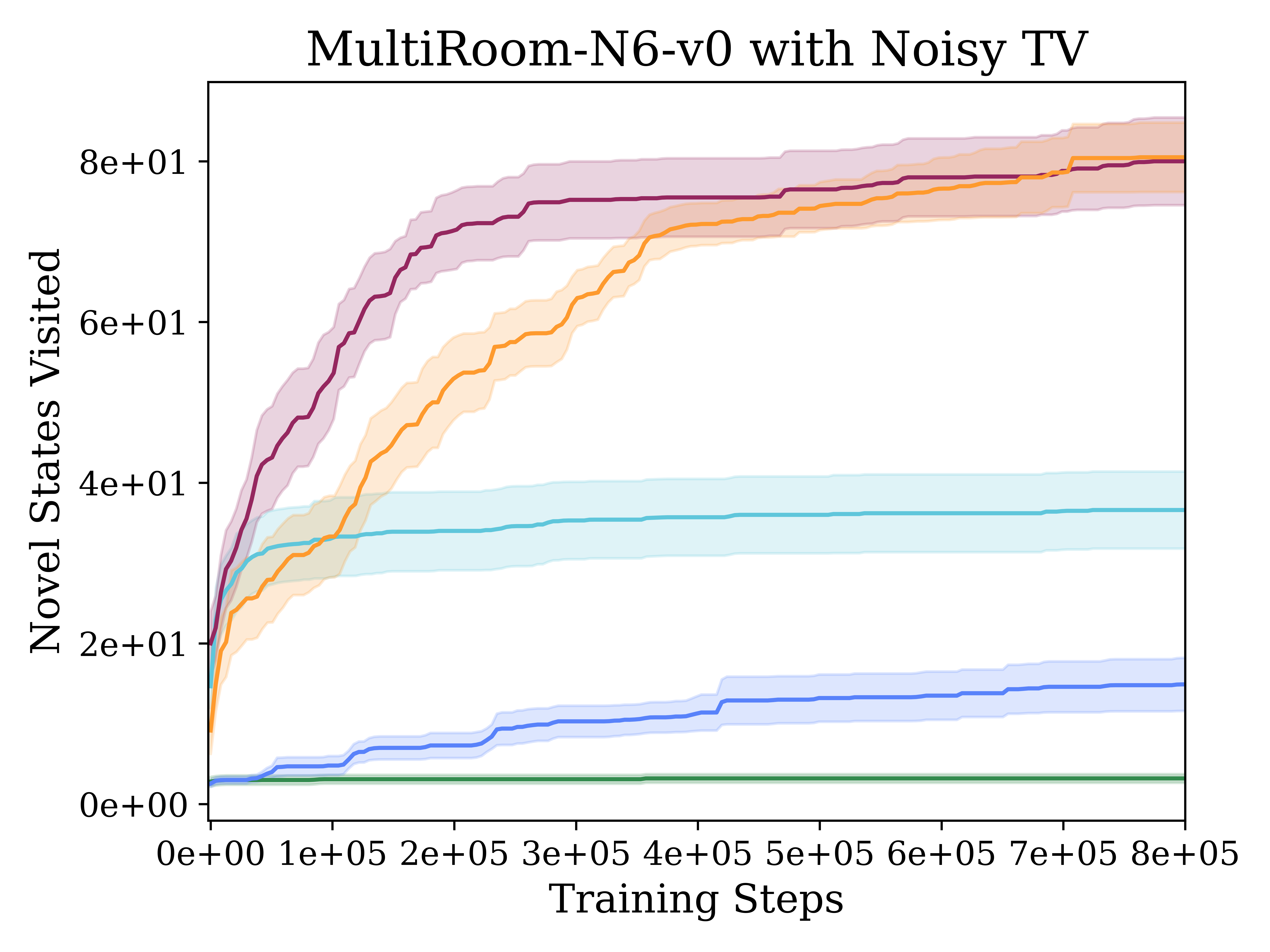}
\caption{} 
\end{subfigure}
\end{minipage}
\hfill
\begin{minipage}{0.35\textwidth}
    \centering 
    \begin{subfigure}{\textwidth}
        \centering 
        \includegraphics[width=0.9\linewidth]{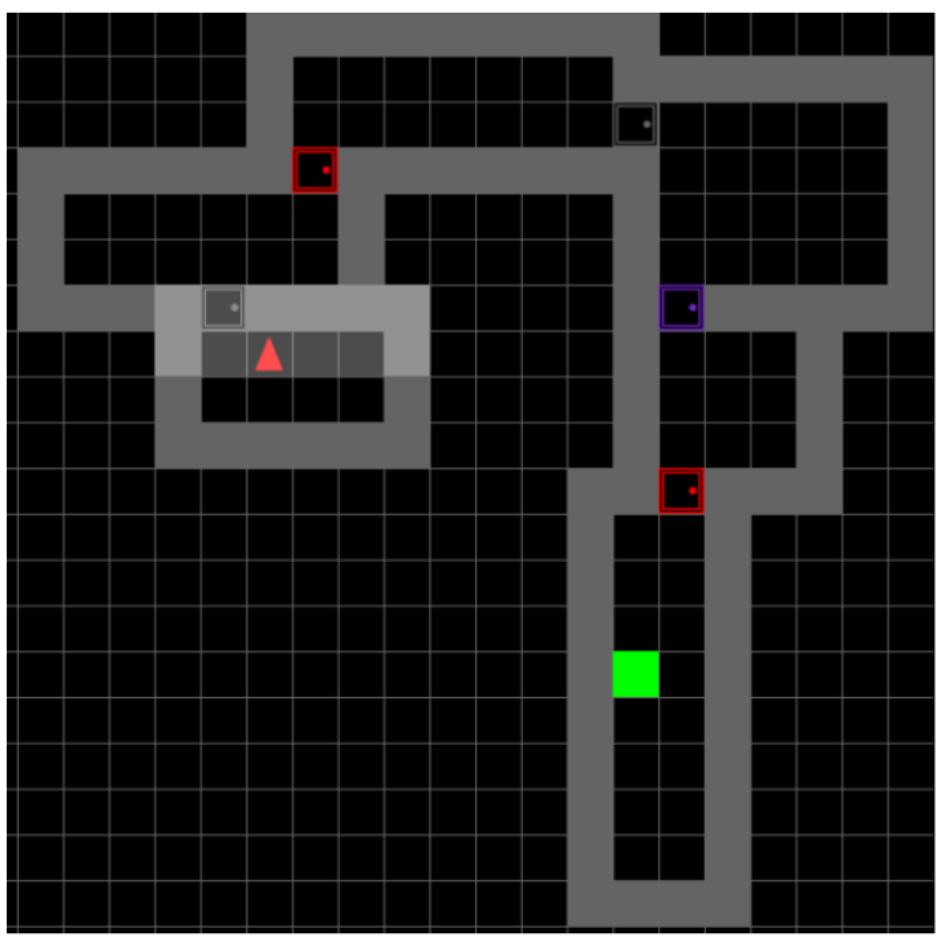}
        \caption{}
    \end{subfigure}
\end{minipage}
\caption{AMA agents effectively explore sparse reward minigrid 
        environments that contain action dependent stochastic
        traps. (a) and (b) panel show
        performance on the easiest
        environment, containing four
        rooms, while the 
        (c) and (d) show performance on
        a more challenging
        environment with six rooms.
        AMA and MSE have similar exploration 
        performance when the noisy TV is absent, 
        outperforming a no-intrinsic-reward baseline---but
        when a noisy TV
        is present only the AMA
        curiosity approach 
        is able to significantly explore the
        environment. Ensemble methods are robust to noisy TVs in this case, but not random network distillation. Panel (e) shows an example six room environment. Standard error 
        represents seed variation. See appendix \ref{sec:minigrid_extrinsic_rewards} for plots of extrinsic rewards.}
    \label{fig:minigrid}
\end{figure*}
The MSE prediction network is unable to
reduce prediction
errors
for the stochastic transitions, causing
it to produce much larger intrinsic
rewards than the
deterministic
transitions, consistent with
\citet{pathak19disagreement}. On
the
other hand, the AMA prediction network is
able to cut its
losses
by
attributing high variance to the
stochastic transitions,
making
them just as rewarding as the
deterministic transitions.

\subsection{Minigrid}
Next we test AMAs on the Gym MiniGrid environment
\cite{gym_minigrid}, which allows for 
resource limited
deep reinforcement learning. The
agent receives 
tensor observations describing its
receptive field at each 
timestep. The channels of the observations
represent
semantic
features (e.g. blue door, grey wall,
empty, etc.) of each grid
tile. The action space is discrete
(containing actions: turn left, turn right, move forward, 
pick up, drop, toggle objects and done) allowing the
agent to move around the environment as wells well as open
and close doors. We used singleton
environments but with different seeds
for each run---resulting 
in different environment configurations for each seed.
We measure exploration by counting
the number of unique
states visited throughout training by the zeroth actor in a distributed RL agent. An
action-dependent noisy TV
was
added,
inspired by other minigrid experiments
with noisy TVs from
\citep{Raileanu2020RIDE:},
by setting approximately
half of the state
observation to 
uniformly sampled integers within the range 
of possible minigrid values. When the
agent selects the `done' action the noisy
TV is activated in
the
next observation. This is the only effect of the 
`done' action. 
\begin{figure*}[tp]\label{fig:retro_games}
     \centering
     \begin{subfigure}[b]{0.24\textwidth}
         \centering
         \includegraphics[width=\textwidth]{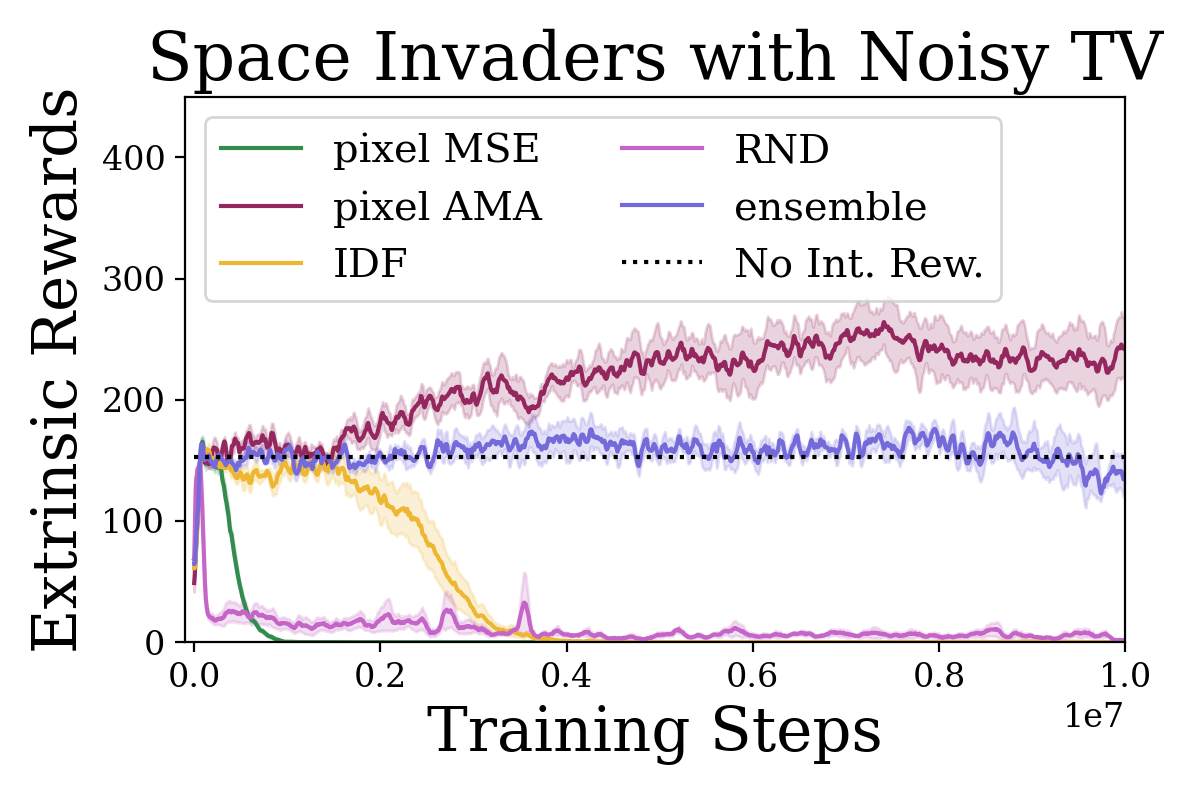}
         \caption{}
     \end{subfigure}
     \hfill
     \begin{subfigure}[b]{0.24\textwidth}
         \centering
         \includegraphics[width=\textwidth]{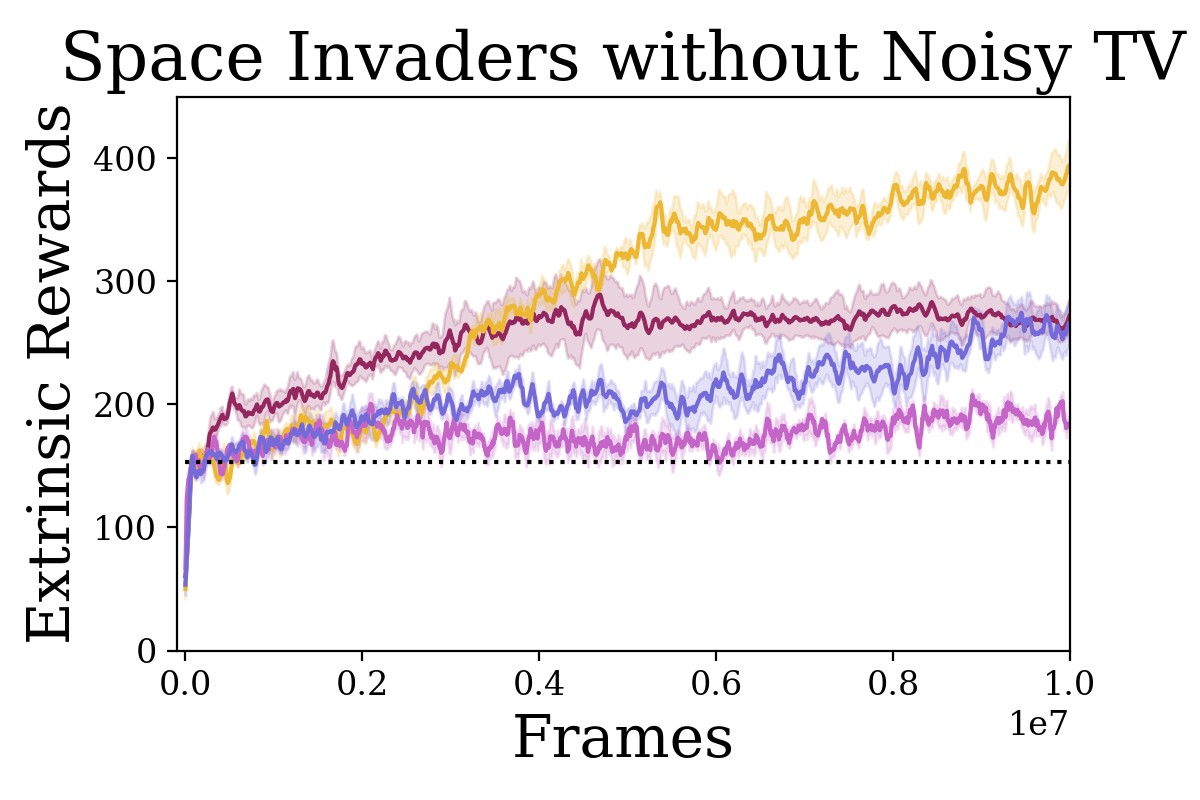}
         \caption{}
     \end{subfigure}
     \hfill
     \begin{subfigure}[b]{0.24\textwidth}
         \centering
         \includegraphics[width=\textwidth]{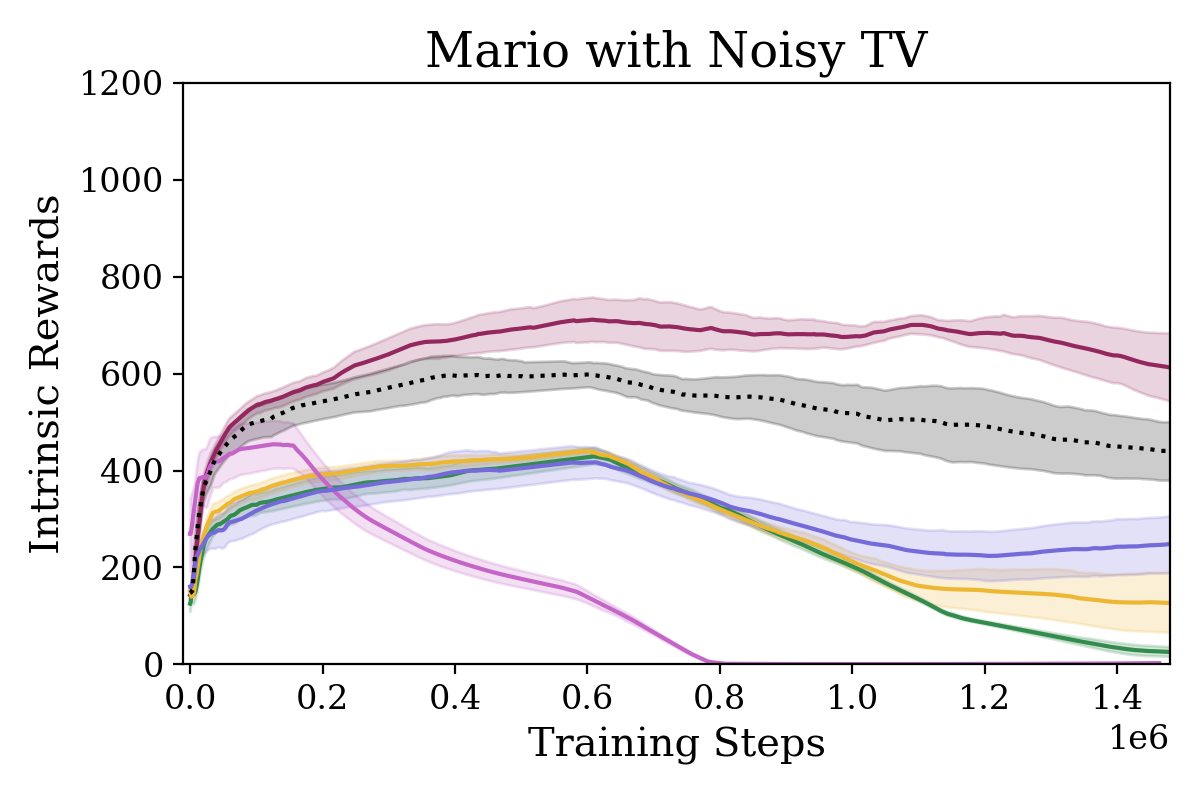}
         \caption{}
     \end{subfigure}
     \hfill
     \begin{subfigure}[b]{0.24\textwidth}
         \centering
         \includegraphics[width=\textwidth]{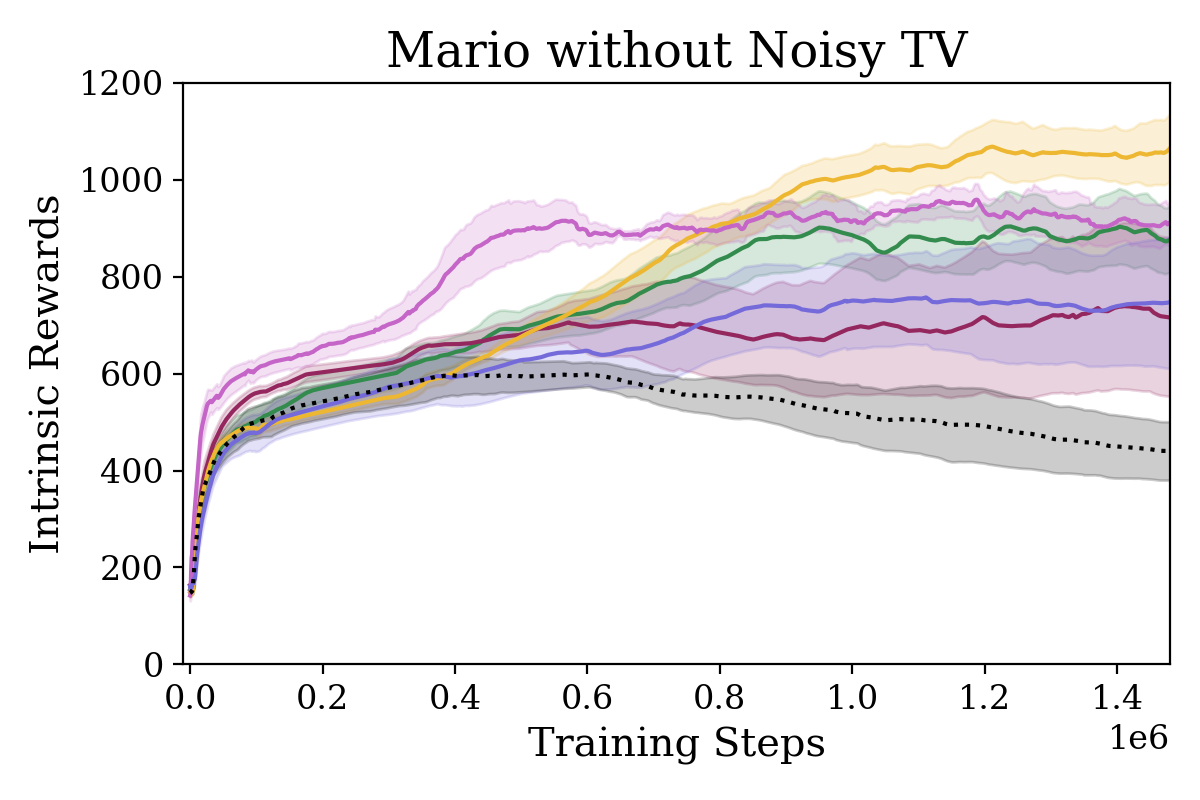}
         \caption{}
     \end{subfigure}
        \caption{ 
        Pixel AMA performs
        significantly better than
        all 
        baselines with a noisy TV (a) and (c) and
        without the distracting noisy TV AMA nearly matches its most
        directly comparable
        method Pixel MSE (b) and (d).
        No extrinsic rewards
        were used for policy
        optimisation. In Mario distance covered and extrinsic reward are equivalent. The y-axis plots extrinsic rewards per episode. The no intrinsic reward baseline for Space Invaders is the mean performance of the random agent data provided by \citet{burda2018large}. For Mario we were required to compute the no intrinsic reward baseline from scratch, which is the PPO agent from \citet{burda2018exploration} with intrinsic rewards turned off \textit{without} a noisy TV (but repeated in all panels for easy comparison).}
        \label{fig:retro_games}
\end{figure*}
We perform policy optimisation with a
synchronous advantage
actor
critic
(A2C) implementation recommended by the
gym minigrid README 
\citep{mnih2016asynchronous,
rl_starter_files}. For the minigrid experiments 
we train on intrinsic and
extrinsic rewards with their relative
weighting being equal. The actor critic
weights were optimised with RMSProp
\citep{tieleman2014rmsprop}
at
a learning rate of 0.001, while the
intrinsic reward module was
optimised with the Adam \citep{kingma2014adam} 
optimizer at a learning rate of 0.001
for
the AMA agent and 0.0001 for the MSE agent. All methods used the same A2C base implementation \citep{rl_starter_files} with the same default hyperparameters. Intrinsic reward specific hyperparameters were optimised for each baseline individually 
with and without the noisy TV, see Appendix \ref{minigrid_hyperparams} for details.

The uncertainty budget $\lambda$ 
of the AMA network was set to 0.1 as the environment representations from minigrid are very sparse, which we found empirically reduces the prediction networks willingness to predict uncertainties. The uncertainty weighting $\eta$ was set to 1 (its natural value). We found that clipping intrinsic rewards to the range $[0, \infty]$ compensated for possible over predictions of uncertainty, which we implemented for AMA. The robustness of performance to hyperparameters is analysed in Section \ref{hyperparam_ablations}.

The forward
prediction module of AMA works in
the
observation space
as
opposed to a learned feature space as is
implemented in other
curiosity driven methods
\citep{pathak2017curiosity,
burda2018exploration}.
Pixel based curiosity
was chosen due to its simplicity. The forward
prediction model is a double headed CNN, which builds upon a previous 
intrinsic motivation implementation on minigrid \citep{Raileanu2020RIDE:}. The ensembles' intrinsic rewards were implemented based on \citet{pathak19disagreement} using the same forward prediction network as AMA. The RND approach was adapted from \citet{Raileanu2020RIDE:}'s implementation.

We perform experiments in four and six
room 
configurations of
the 
minigrid (see Figure \ref{fig:minigrid}(e) for an example six room environment). Without a noisy TV both AMA and MSE reward functions generate 
visit more states compared to the no intrinsic reward baseline. 
On the other hand, the presence of a noisy TV profoundly affects the performance
of the MSE curiosity agent, greatly reducing the number of states
visited. In contrast, AMA agents are almost unaffected by the presence 
of an action dependent noisy TV. RND \cite{burda2018exploration} shows good performance without a noisy TV but performs poorly when a noisy TV is present. Ensemble disagreement \cite{pathak19disagreement} demonstrate robustness to this version of a noisy TV in minigrid---preserving its exploration boost over A2C with and without the TV.
\subsection{Mario and Space Invaders}
We have shown AMAs can learn to ignore stochasticity in the supervised setting of MNIST and also explore proficiently in the sparse reward multiroom environments of minigrid. In this section we test whether AMAs can learn to explore in stochastic versions of high dimensional retro video game environments. We isolated two games that have been used as curiosity benchmarks in the relevant exploration literature: Space Invaders (from \citet{bellemare2013arcade} used in \citet{burda2018large, pathak19disagreement}) and Mario (from \citet{nichol2018retro} used in \citet{pathak2017curiosity, burda2018exploration, pathak19disagreement}).

While these environments are useful for testing curiosity, they are mostly deterministic. Sticky action environments \cite{machado2018revisiting} were previously developed to make atari RL algorithms more robust to stochasticity. However, the stochasticity in sticky action environments is independent of the agents policy, meaning an agent cannot trap itself by selecting actions that generate random dynamics. This is unlike random dynamics that are likely to be found in real world applications---for example a curious warehouse robot could become distracted with watching another robots actions whose goal it does not understand. Consequently, we developed a noisy TV wrapper for atari game environments, where the action space is extended with an action that induces random grayscale tiled CIFAR-10 \citep{krizhevsky2014cifar} images in place of game frames for the next observation (using\footnote{\url{https://github.com/snatch59/load-cifar-10}}, see Appendix \ref{CIFAR} for examples). When the noisy TV action is selected, the zeroth action of the action space is sent to the game emulator (the choice of the zeroth action is arbitrary). 

We adapt the proximal policy optimisation (PPO) \citep{schulman2017proximal}
curiosity implementation from \citet{burda2018large} into an AMA curiosity system.
When using the pixel feature space, we extend \citet{burda2018large}'s U-Net \citep{ronneberger2015u} to use two output heads to predict the mean and variance of future states. We leave all PPO hyperparameters equal to their values from \citet{burda2018large}. We set the uncertainty budget hyperparameter $\lambda$ to 1 and the uncertainty weighting hyperparameter $\eta$ to 2---doubling the punishment given to our agents for experiencing aleatoric uncertainty. For the these experiments we do not clip intrinsic rewards. The hyperparameter optimisation process is described in Appendix \ref{retro_hyperparameters}. We also test the robustness of AMA to alternative noise distributions (Section \ref{uniform_noise}) as well as different settings of hyperparameters (Section \ref{hyperparam_ablations}). We compare AMA to four alternative 
intrinsic reward methods: random
network distillation (RND) 
\citep{burda2018exploration}, inverse
dynamics feature (IDF) curiosity 
\citep{pathak2017curiosity}, MSE pixel
based curiosity \citep{burda2018large} and 
ensemble disagreement \citep{pathak19disagreement}. It is important to stress \textit{policy
optimisation is done with
intrinsic 
rewards only} following \citet{burda2018large}. 

Compared to the
relatively 
weak baseline of pixel based
curiosity, 
the Space Invaders and Mario experiments
show similar results to the 
minigrid experiments---MSE 
and AMA pixel based curiosity have
comparable performance when no
noisy TV 
is present (Figure \ref{fig:retro_games}(b) and \ref{fig:retro_games}(d)), 
while with the noisy TV AMA
greatly outperforms
MSE pixel based curiosity (Figure \ref{fig:retro_games}(a) and 
Figure \ref{fig:retro_games}(c)).
Unsurprisingly, RND and IDF curiosity 
maintain their superiority over pixel
based methods
without a noisy TV (Figure \ref{fig:retro_games}(b) and \ref{fig:retro_games}(d)).
However, unlike AMA
curiosity, both 
these baselines are vulnerable to 
action dependent noisy TVs (Figure \ref{fig:retro_games}(a) and 
Figure \ref{fig:retro_games}(c)).
While Ensembles and RND have previously been shown to be able to evade stochastic traps 
\cite{burda2018exploration}, it seems that if the number of novel states 
in the trap is large then their approximation of epistemic uncertainty breaks down. 
\begin{figure}
     \centering
     \begin{subfigure}[b]{0.49\columnwidth}
         \centering
         \includegraphics[width=\textwidth]{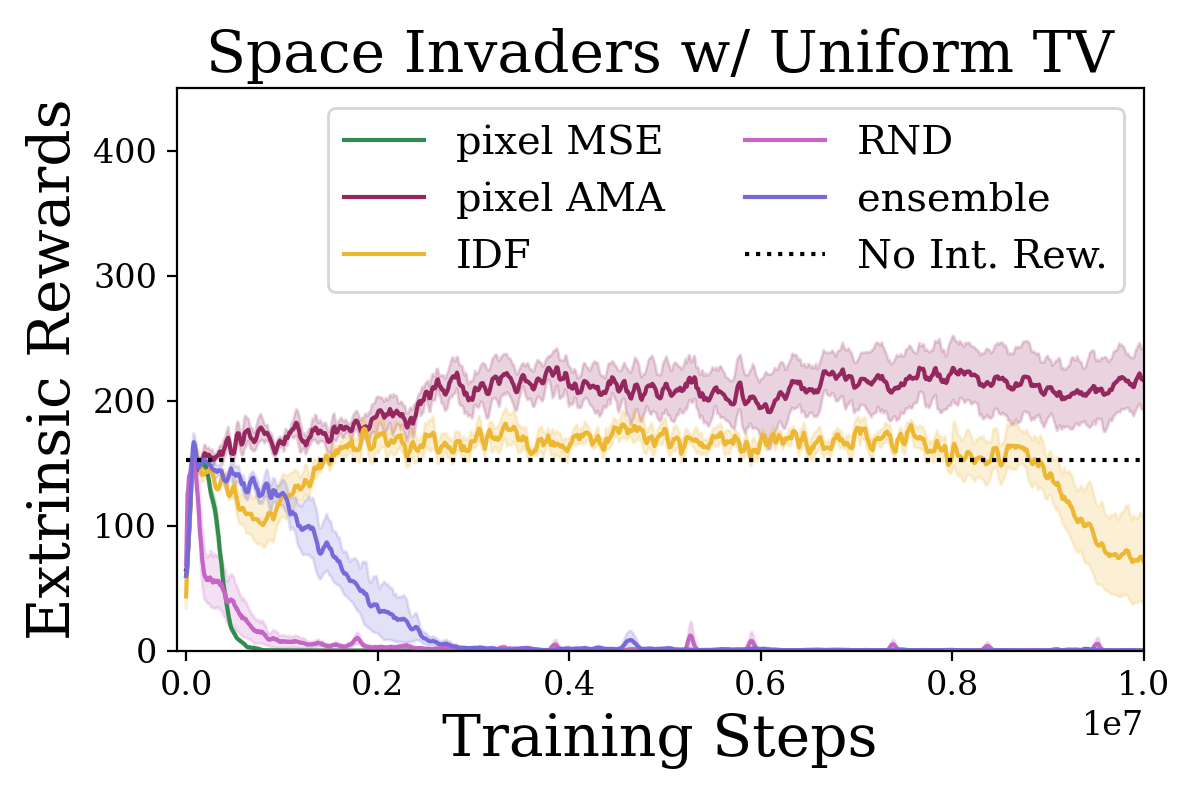}
         \caption{}
     \end{subfigure}
     \begin{subfigure}[b]{0.49\columnwidth}
         \centering
         \includegraphics[width=\textwidth]{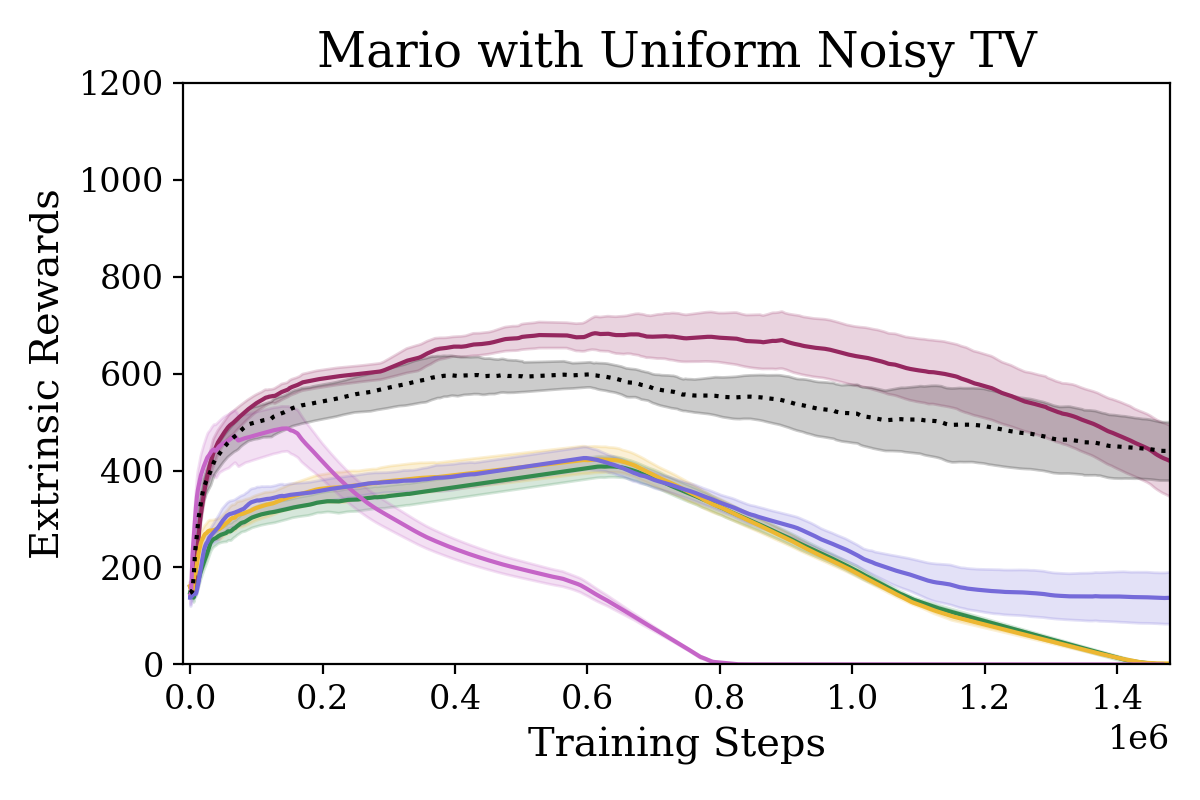}
         \caption{}
     \end{subfigure}
     \caption{AMA is robust to Noisy TVs of a very random unifrom noise distribution 
    (random pixels from 0-255) while other baselines are also trapped by this additional noisy TV. We verify in \ref{mario} the eventual collapse in Mario is a property of the curiosity methods tested and not a feature of AMA.}
    \label{fig:salt_and_pepper}
\end{figure}
\subsection{Bank Heist: An Atari Game with a Natural Noisy TV}\label{bank_heist}

Although Atari games are famously deterministic \cite{machado2018revisiting}, we identified 
a naturally ocurring stochastic trap in the Bank Heist gameplay videos of the original IDF curiosity paper \cite{pathak2017curiosity}. 
The objective of Bank Heist is to
simultaneously 
avoid police cars and navigate to banks 
distributed across four 2D mazes---which can 
be entered and exited through the sides of the
screen. Importantly, with each enter/exit 
the bank locations reset randomly. 

When trained on purely intrinsic rewards, 
IDF curiosity will perpetually enter 
and exit the maze while also dropping
dynamite. This can create high 
prediction error as it is impossible to
predict when the 
dynamite will explode and where the banks
will regenerate. An example video of this behaviour orignally from \cite{pathak2017curiosity} is provided\footnote{\url{https://www.youtube.com/watch?v=S4YdZe70XMQ}}. To measure the effect of this pathological behaviour on exploration, we count the number of pixels covered by the car on average in an episode. 

\begin{figure}\label{bank_heist_figure}
     \centering
     \begin{subfigure}[b]{0.49\columnwidth}
         \centering
         \includegraphics[width=\textwidth]{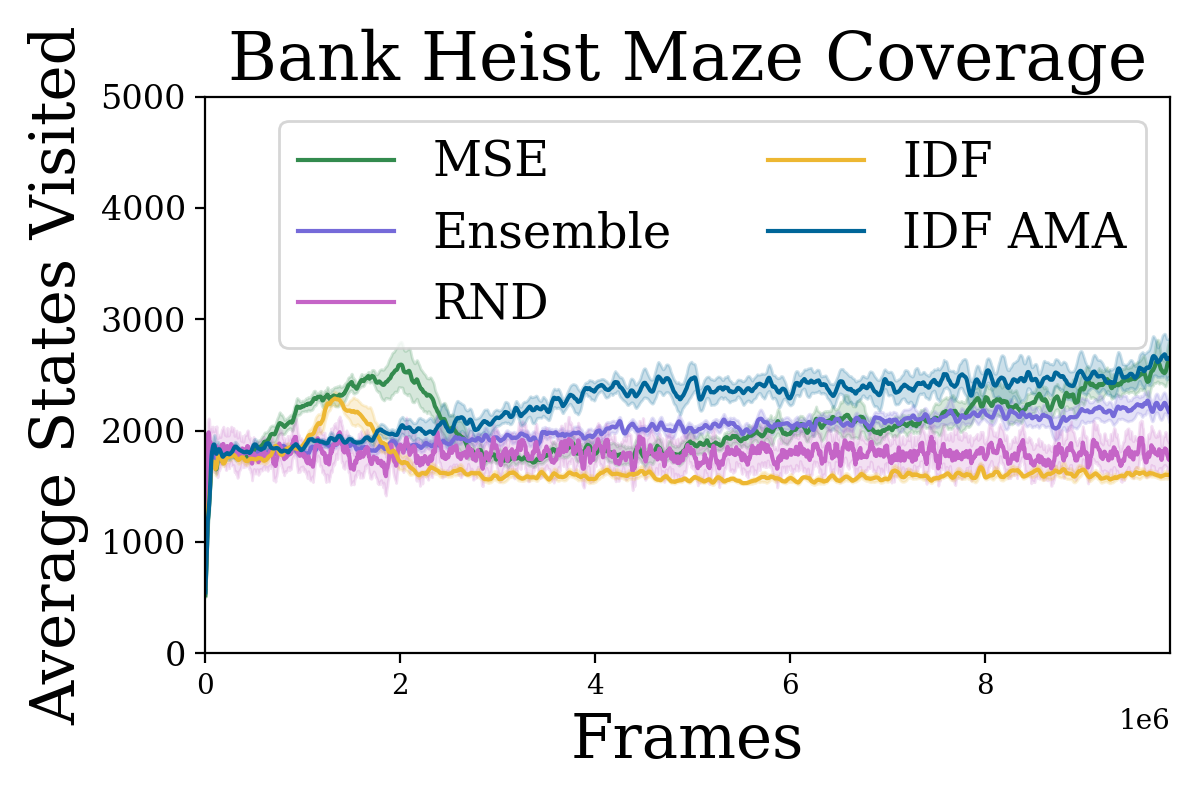}
         \caption{}
     \end{subfigure}
     \begin{subfigure}[b]{0.49\columnwidth}
         \centering
         \includegraphics[width=\textwidth]{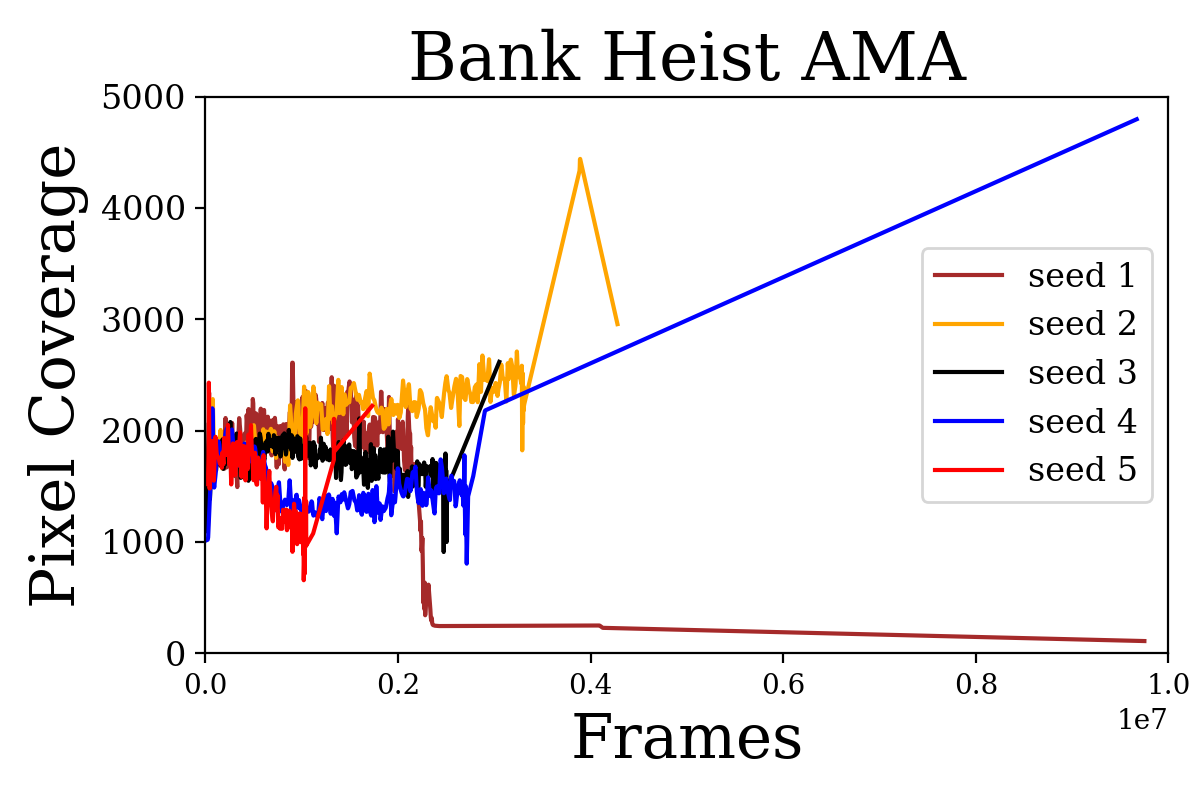}
         \caption{}
     \end{subfigure}
     \caption{IDF AMA (a) and Pixel AMA (b) avoids the
        natural trap in Bank Heist
        and so explores more of the maze on
        average than the IDF
        MSE. (b) shows the different seeds from Pixel AMA plotted individually due to the very long episodes of pixel AMA. An implementation detail means the values of the x-axis above are a close approximation to their true values see \ref{bank_heist_x_explanation} for details.}
        \label{fig:bank_heist}
\end{figure}

As expected, the IDF agent initially begins to explore the environment before coverage decreases (Figure \ref{fig:bank_heist}(a)). On the other hand, most of the other baselines seem to be relatively unaffected by this trap---although they do not seem to explore particularly proficiently. Pixel AMA seems to develop an interesting strategy which makes its performance difficult to plot---so we plot all of its seeds individually in Figure \ref{fig:bank_heist}(b). Pixel AMA discovers does not rob any banks so that it is not be chased by the police, allowing it to explore unimpeded with very long episodes (so long that they exceed the $10^{7}$ timesteps used for training hence why the line is cut off at different points for different seeds). Overall, pixel AMA explores the maze well, with 2 seeds getting good performance, 2 seeds getting extremely good performance and 1 seed becoming bored and deciding to give up on exploration. Lastly, we integrated the AMA prediction paradigm into the IDF
approach, predicting the mean and aleatoric variance of future state
representations and computing rewards with Equation
(\ref{eqn:intrinsic_reward_function}) with AMA hyperparameters 
$\lambda$ and $\eta$ set to 1. This ablation was not susceptible to the natural Bank Heist trap (Figure \ref{fig:bank_heist}). Overall, this section highlights that noisy TVs can be 
an inherent issue even in environments with simple dynamics. 

\subsection{Different Noise Distributions}\label{uniform_noise}
A key desiderata of AMA is that exploration performance should be robust to noisy TVs no matter the distribution they follow. In addition to the CIFAR Noisy TV experiments, we performed further experiments to see how AMA and the baselines would react to uniform ``salt and pepper'' noise in Mario. Although uniform noise is a simpler distribution than natural images, it is even more distracting than CIFAR images due to its high entropy. Consequently, AMA is again the only method that does not catastrophically fail when a noisy TV is present as can be seen in Figure \ref{fig:salt_and_pepper}. Interestingly, ensembles perform considerably worse with uniform noise than with CIFAR noise. AMA's extrinsic reward decreases slightly in Mario---we verified in appendix \ref{mario_decreasing} that this is a result of the well known boredom behaviour of curiosity systems discovered by \cite{pathak2017curiosity}, rather than being distracted by a Noisy TV.
\subsection{Hyperparameter Ablations}\label{hyperparam_ablations}
The hyperparameters $\lambda$ and $\eta$ have simple interpretations---$\lambda$ is a prior on the width of the noise distribution, while $\eta$ controls how much AMA is punished for experiencing random dynamics. Although $\lambda$ and $\eta$ are interpretable, it is important to understand the effect on exploration performance when hyperparameters are perturbed from their optimal values. Table 1 shows that Mario performance in relatively robust to hyperparameter selection but as expected $\eta$ and $\lambda$ can be tuned to be more susceptible to noisy TVs. Table 2 shows that it is not crucial to tune $\eta$ minigrid but due to the nature of minigrid observations it is important to have a wide prior on the variance of states. An interesting direction for further work would be to try to estimate $\lambda$ online from experience. 
\section{Discussion}
\subsection{Limitations}
The AMA reward function implicitly rewards epistemic 
uncertainty by assuming the total uncertainty can be decomposed 
into epistemic and aleatoric uncertainties. While theoretically true
\cite{kendall2017uncertainties}, 
there is no guarantee that AMAs are able to surgically subtract those 
errors due to aleatoric dynamics from the total prediction error. 
Additionally, aleatoric 
uncertainty estimates are not guaranteed to be reliable for out 
of distribution data, meaning intrinsic rewards could become less reliable 
the further the agent travels into novel territory \cite{mukhoti2021deterministic}. 
In practice we find that a stochastic policy---and clipping in the case of
minigrid---offsets 
potentially deceptive intrinsic rewards. Furthermore, trust region methods in 
the retro games (i.e. using PPO \cite{schulman2017proximal} 
instead of A2C \cite{mnih2016asynchronous}) may also compensate for 
ocassionally deceptive rewards---suggested by the fact 
that intrinsic reward clipping was not necessary for the retro game
experiments. 
Finally, we note that like all curiosity approaches (e.g.
\citet{burda2018large}),
our method generates non-stationary rewards, 
which is known to make learning difficult for RL agents.

\subsection{Acetylcholine}\label{ACh}
AMAs build upon \cite{yu2003expected}'s work on Acetylcholine---a neurotransmitter associated with ``expected'' uncertainty signalling in the brain (we point the interested reader to Appendix \ref{ACh_review} for a literature review). The exact nature of the uncertainty signalled by acetylcholine---whether it is epistemic or aleatoric---is an open question in theoretical neuroscience posed by \cite{angela2005uncertainty}. Previously, it was obvious why a biological agent would find epistemic uncertainty predictions useful as they can be used to maximise information gain \cite{mukhoti2021deterministic}. In this work, we present a use case for aleatoric uncertainty predictions---rejuvenating interest in \cite{angela2005uncertainty}'s call for experiments to analyse the kinds of uncertainty signalled in brain. To codify the relevance of AMAs to theoretical neuroscience, we propose an animal experiment in \ref{bandit} as well as theoretical predictions of cholinergic activity in the proposed task. 
\begin{table}
\centering
\begin{tabular}{ |c|c|c|c|c| } 
\hline
$\eta$ & $\lambda$ & Max. X Distance & Max. X Distance w/ TV \\
\hline
2 & 0.1 & 598 & 641\\ 
2 & 1 & 616 & 715 \\ 
2 & 10 & 1092 & 451 \\ 
1 & 1 & 725 & 814 \\ 
0.5 & 1 & 568 & 516 \\ 
\hline
\end{tabular}
\caption{Mario results with different hyperparameters. Hyperparameter tuning is not crucial in Mario. However, increasing $\lambda$ or decreasing $\eta$ can make AMA susceptible to noisy TVs.}
\end{table}\label{table:mario_table}

\begin{table}
\centering
\begin{tabular}{ |c|c|c|c|c| } 
\hline
$\eta$ & $\lambda$ & Novel States & Novel States w/ TV \\
\hline
1 & 0.1 & 100 & 92 \\ 
1 & 1 & 85 & 5 \\ 
1 & 10 & 44 & 3 \\
2 & 0.1 & 115 & 107\\
0.5 & 0.1 & 100 & 70 \\ 

\hline
\end{tabular}
\caption{Minigrid results with different hyperparameters. Due to the sparse observations of minigrid, it is important to place a prior of being more eager to predict uncertainty by decreasing $\lambda$. For minigrid, tuning $\eta$ is not very important.}
\end{table}\label{minigrid_table}

\section{Conclusion}
We have shown AMAs are able to avoid
action-dependent
stochastic
traps that destroy the exploration
capabilities of
conventional
curiosity driven agents in environments with high entropy noisy TVs
\cite{burda2018large, pathak19disagreement, burda2018exploration}. 
AMAs tractably avoid stochastic traps 
by decreasing intrinsic
rewards in regions with high estimated
aleatory. Future RL research should aim to
integrate the AMA approach into curiosity methods that operate 
in feature spaces besides pixels or 
even within those methods that circumvent dynamics 
altogether (e.g. \citet{burda2018exploration}), with the aim of 
achieving state of the exploration even when noisy TVs are present. Lastly, more work should be done to understand the cause of the failure cases of popular intrinsic reward methods in stochastic environments. 
\section*{Acknowledgements}
Augustine N. Mavor-Parker is supported by the EPSRC project reference 2250955. Caswell Barry was funded by a Wellcome Senior Research Fellowship (212281/Z/18/Z). Augustine N. Mavor-Parker would like to thank Changmin Yu, Andrea Banino, Charles Blundell, Maximillian Mozes, Kimberly Mai and Felix Biggs for their comments on this project.
\bibliography{main}
\bibliographystyle{icml2021}
\appendix
\onecolumn

\section*{Appendix}
\section{Acetylcholine}\label{ACh_review}
In the mammalian brain acetylcholine is 
implicated in a range of processes
including learning and
memory, fear, novelty detection, and
attention
\cite{ranganath2003neural,
    pepeu2004changes,
acquas1996conditioned, barry2012possible,
angela2005uncertainty,
hasselmo2006role, giovannini2001effects,
parikh2007prefrontal}.
Traditional views---supported by the
rapid increase in
cholinergic tone in response to environmental
novelty and
demonstrable
effects on neural plasticity---
emphasised its role as a learning signal,
generating physiological
changes that favour encoding of new
information over retrieval
\cite{hasselmo2006role}.

Notably, \cite{yu2003expected} proposed an alternative perspective,
suggesting that acetylcholine signals the {\it expected uncertainty}
of top
down
{\it predictions}, while modulation of norepinephrine is a
result of 
{\it unexpected uncertainties}. 
More concretelty, \cite{yu2003expected}'s model
can be seen as favouring bottom up 
sensory input over top down predictions if 
predictions are believed to be inaccurate---consistent 
with evidence that shows acetylcholine
inhibits feedback connections and strengthens sensory inputs
\cite{hasselmo2006role}.
However this approach does not explicitly
separate epistemic 
and aleatoric
uncertainties \cite{angela2005uncertainty}. In contrast, the
utility of
quantifying epistemic
uncertainties for exploration has been widely recognised in the
RL
literature
(e.g. \cite{osband2016deep, pathak19disagreement}). 
Here we demonstrate a potential use of aleatoric uncertainties
in exploring agents both biological and artificial. Namely,
aleatoric
uncertainties can be used to divert attention away from
unpredictable dynamics
when using prediction errors as intrinsic rewards. 
This is similar to a model proposed by \cite{parr2017uncertainty}, suggesting 
acetylcholine may indicate expected uncertainties in top down
predictions within an MDP.

In this context we propose an extension to \cite{angela2005uncertainty}’s
dichotomy.
Specifically, we suggest that in the mammalian brain, cortical
acetylcholine
signals expected aleatoric uncertainties, while norepinephrine
is
modulated by
epistemic uncertainties both expected and unexpected. This
formulation is
attractive in an ML framework, providing a means to avoid
stochastic traps,
while also being consistent with empirical biological data \cite{hasselmo2006role, yu2003expected}.

\section{A Proposed Test for the Aleatoric Model of Acetylcholine}\label{bandit}
Inspired by \cite{angela2005uncertainty}, 
we propose that in the mammalian brain acetylcholine signals aleatoric uncertainty surrounding
future states. However, we are not aware of any experimental neuroscience data that 
elucidates the specific nature of the uncertainty signalled by acetylcholine. As a result, 
this section proposes a 1D rodent VR task 
designed to test the specific nature of cholinergic uncertainty signalling in 
the mammalian brain which we hope will be picked up by experimental neuroscientists. 
To supplement our experimental proposal, 
we compute theoretical predictions of cholinergic activity within either an 
aleatoric or epistemic acetylcholine model---two competing interpretations of 
\cite{angela2005uncertainty}'s work. The aleatoric model uses aleatoric uncertainties as 
a theoretical acetylcholine signal \cite{kendall2017uncertainties}, 
while the epistemic model uses ensemble variance as an acetylcholine signal \cite{pathak19disagreement}. 

The proposed task places an animal 
in a VR corridor containing
a series of spatial landmarks and two reward zones in which it
must respond in order to have a chance of receiving a reward. Responding 
in reward zone A causes the animal to teleport to a random position along 
the track. Responding in reward zone B causes the animal to teleport to 
a fixed position on the track. 

\begin{figure*}[tp]
     \centering
     \begin{subfigure}[b]{0.32\textwidth}
         \centering
         \includegraphics[width=\textwidth]{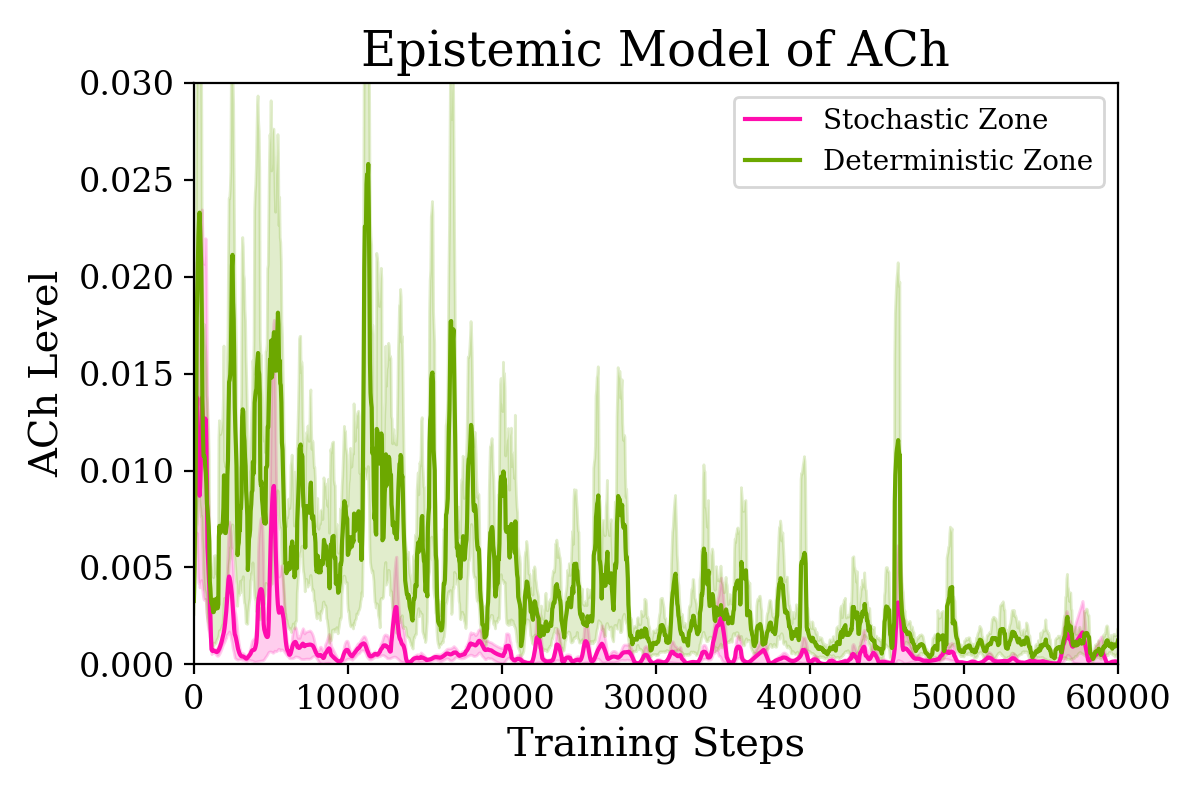}
         \caption{}
     \end{subfigure}
     \hfill
     \begin{subfigure}[b]{0.32\textwidth}
         \centering
         \includegraphics[width=\textwidth]{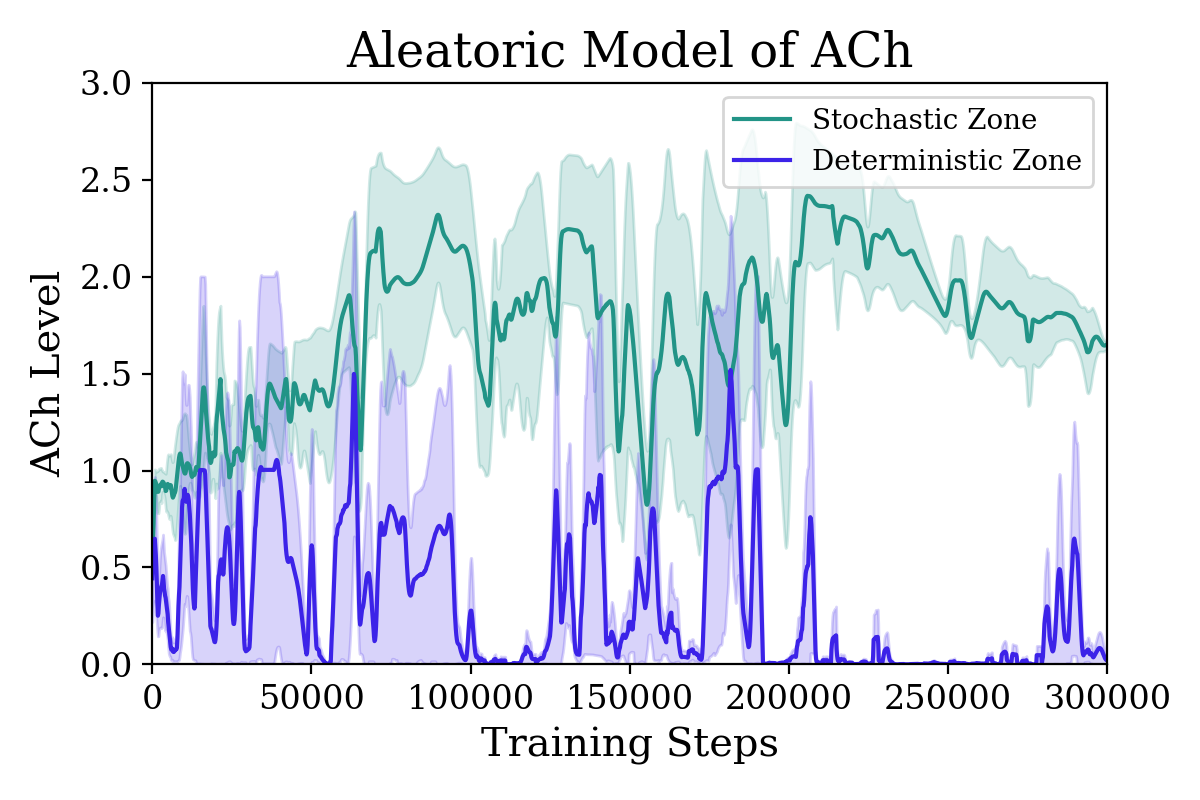}
         \caption{}
     \end{subfigure}
     \hfill
     \begin{subfigure}[b]{0.32\textwidth}
         \centering
         \includegraphics[width=\textwidth]{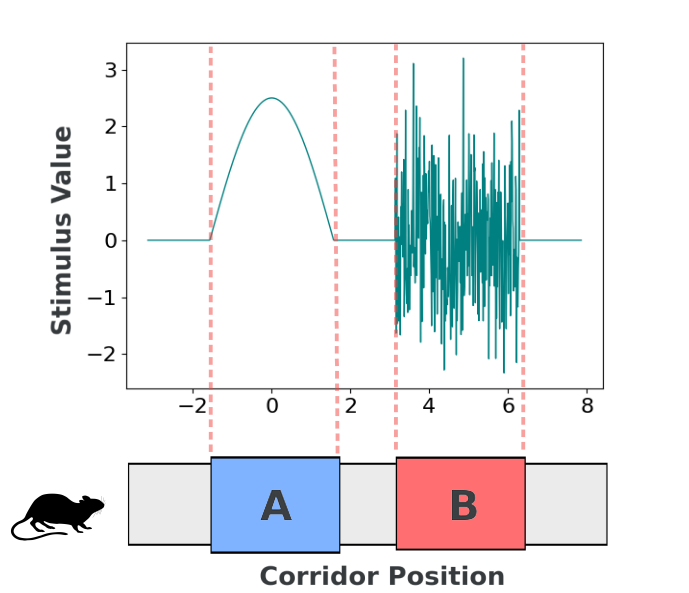}
         \caption{}
     \end{subfigure}
    \caption{Predictions on a theoretical experiment to 
            illuminate the epistemic or aleatoric nature 
            of cholinergic signalling in the brain. (a) and (b) show 
            that the
            epistemic model predicts
            acetycholine will eventually decrease in both zones
            while the aleatoric model predicts acetylcholine decreasing to zero
            in the stable zone but remaining
            high in the noisy zone. Panel (c) shows our 1D model of the 
            proposed animal experiment where the bandit 
            samples different `corridor positions' and receives scalar stimuli, which it is 
            trying to predict---using the resulting prediction errors as
            intrinsic rewards. Plots are smoothed with the same method as the Bank Heist plots \ref{bank_heist_x_explanation}.} 
\end{figure*}

To compute how both models predict the cholinergic signal 
should respond in the proposed rodent VR experiment, we simulate the task with a
simple multi-armed bandit environment. In our bandit model of the task an agent predicts a
1D function by sampling minibatches from different regions of the input. In one
region of sample space the function takes a simple sinuisoidal form, analogous
to zone A of the VR track, in a second region the function consists of points
randomly sampled from a standard normal distribution at each timestep,
analogous to zone B (Figure 6(c)). As described previously, we applied two
models to this task, in the first acetycholine was identified with aleatoric
uncertainty, while in the second---as a control---acetylcholine tracks
epistemic uncertainty. 

We trained an action value based multi-armed
bandit to maximise 
intrinsic rewards for two kinds of forward
prediction models:
a double headed network 
trained to optimise the 
AMA objective and an ensemble of
networks where each 
member is minimising their own MSE (e.g. \cite{pathak19disagreement}). The aleatoric model 
uses the AMA reward function whereas 
the epistemic model is (intrinsically)
rewarded for variance in ensemble predictions.
We plot both models' uncertainties  
in each reward zone over
time---recovering a clear prediction of cholinergic 
activity in both cases.
The aleatoric uncertainty of AMA remains high in reward
zone B but decreases
in
reward
zone A. On the other hand, the epistemic
model shows a decrease in acetylcholine in both
reward
zones over time. We hope these clear and distinct predictions on the nature 
of cholinergic uncertainty signalling will 
be tested by the experimental neuroscience community in a task similar to the one 
we propose.

\section{Implementation Details}\label{implementation_details}
\subsection{Noisy MNIST}
We use three random seeds for the repeats of the MNIST experiments. The results in the graph show test set performance. The hyperparameters used are listed below. The learning rate was manually tuned so that the identity transformation was learned for the deterministic transitions (hence very low loss for the MSE and AMA network) and the AMA network produced sensible uncertainty estimates for the stochastic transitions.

\begin{table}[H]
    \centering
\begin{tabular}{ |p{5cm}|p{5cm}|  }
 \hline
 Hyperparameter & Value\\
 \hline
 MSE Learning Rate & 0.001\\
 AMA Learning Rate & 0.0001\\
 Batch Size & 32\\
 AMA uncertainty budget $\lambda$ & 1 \\
 AMA uncertainty coefficient $\eta$ & 1 \\
 \hline
\end{tabular}\caption{Noisy MNIST hyperparameters}
\end{table}

\begin{figure*}[tb]
    \vspace{-3mm}
    \includegraphics[width=\linewidth]{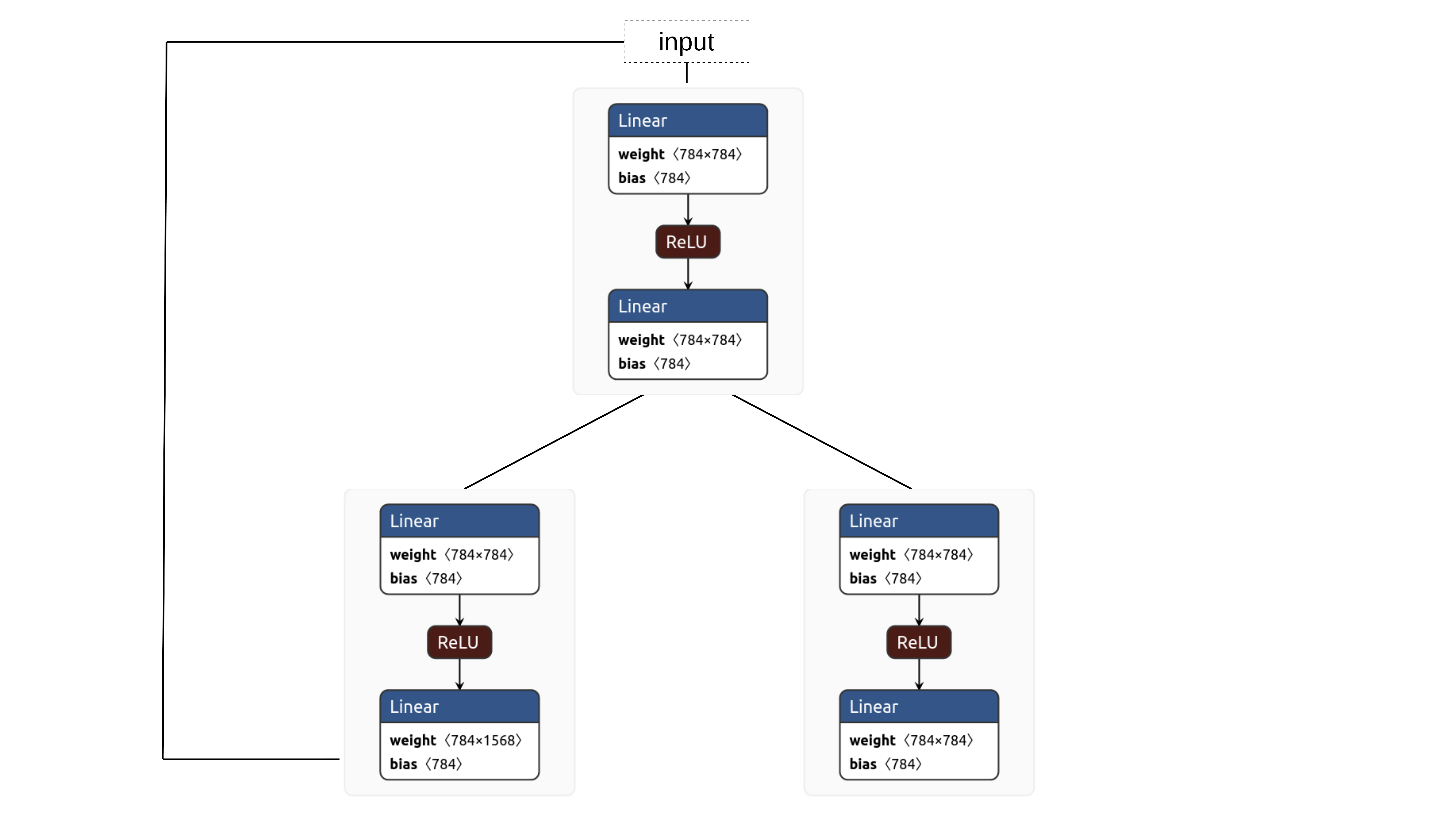}

    \caption{MNIST image to image architecture. For the MSE network the second log variance branch was discarded. A skip connection was provided from the input to the final layer for both AMA and MSE.} 
\end{figure*}

\subsection{Minigrid}\label{minigrid_hyperparams}
We used 5 seeds for the minigrid experiments when tuning the learning rate for respective curiosity modules via grid search: $\in[0.01, 0.001, 0.0001]$, we selected the learning rate for each method that achieved both good exploration (as measured by novel states visited) and extrinsic rewards (if any were achieved) for both the noisy TV and no noisy TV settings on the 6 room environment. We give equal weighting to intrinisc and extrinsic rewards. We used different seeds for the grid search and the final results. For the final results we used 10 seeds. We also adapt an implementation of the Welford algorithm from stack overflow for normalising rewards \footnote{\url{https://stackoverflow.com/a/5544108/13216535}}. We found to return to goal states it is important to normalise the sum of extrinsic and intrinsic rewards, which we now implement for the novel states visited results in the main text (the previous manuscript only normalised intrinsic rewards in the novel states visited results shown in the main text). The architecture for forward prediction is adapted from the implementation from \cite{Raileanu2020RIDE:} but in preliminary experimentation we ended up changing their prediction architecture dramatically. 
\subsection{Extrinsic Rewards Experiments in Minigrid}\label{sec:minigrid_extrinsic_rewards}
\begin{figure*}[tp]
\centering
\begin{minipage}{0.59\textwidth}
\begin{subfigure}{0.49\textwidth}
\includegraphics[width=\linewidth]{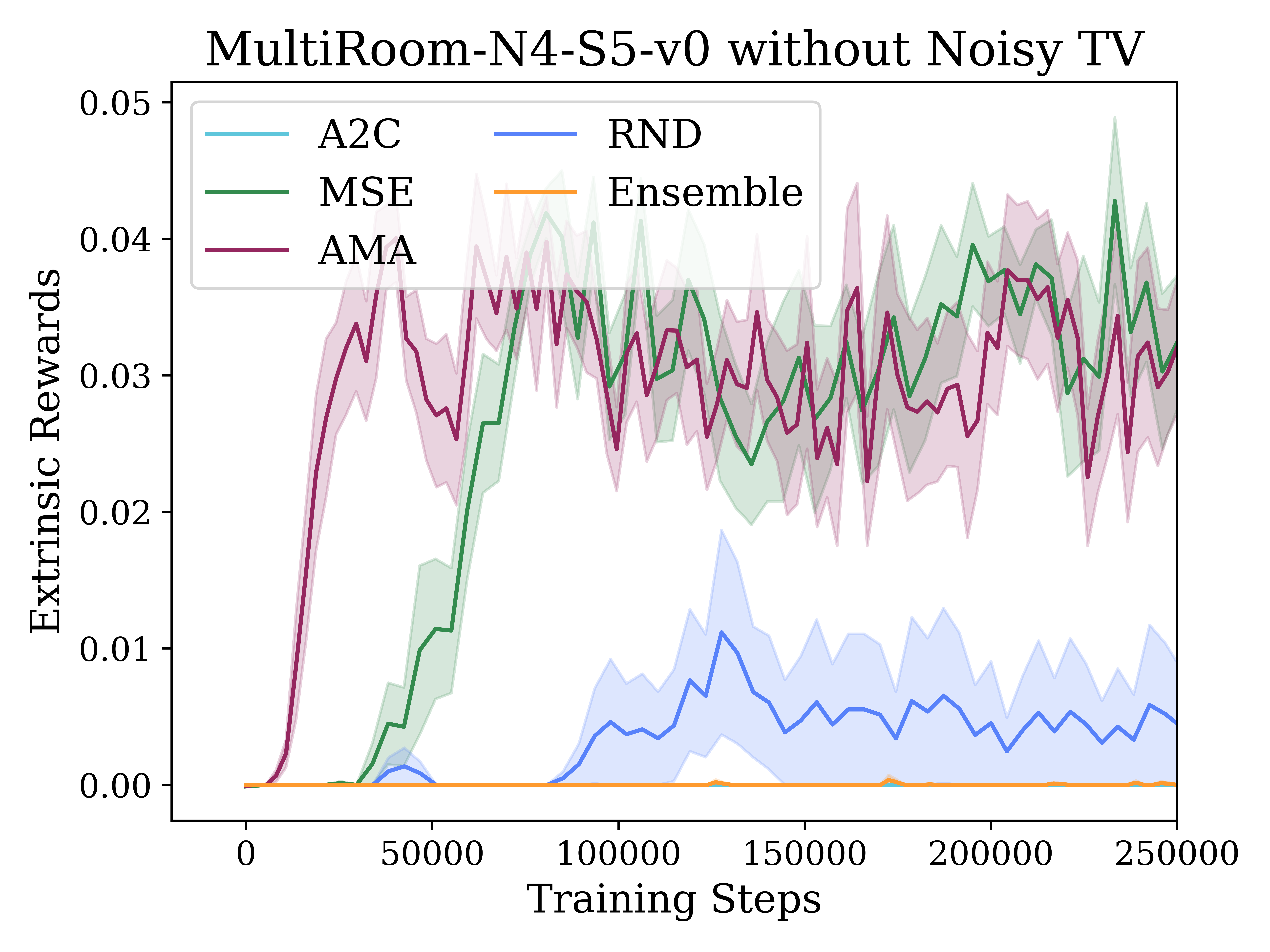}
\caption{} 
\end{subfigure}
\hfill
\begin{subfigure}{0.49\textwidth}
\includegraphics[width=\linewidth]{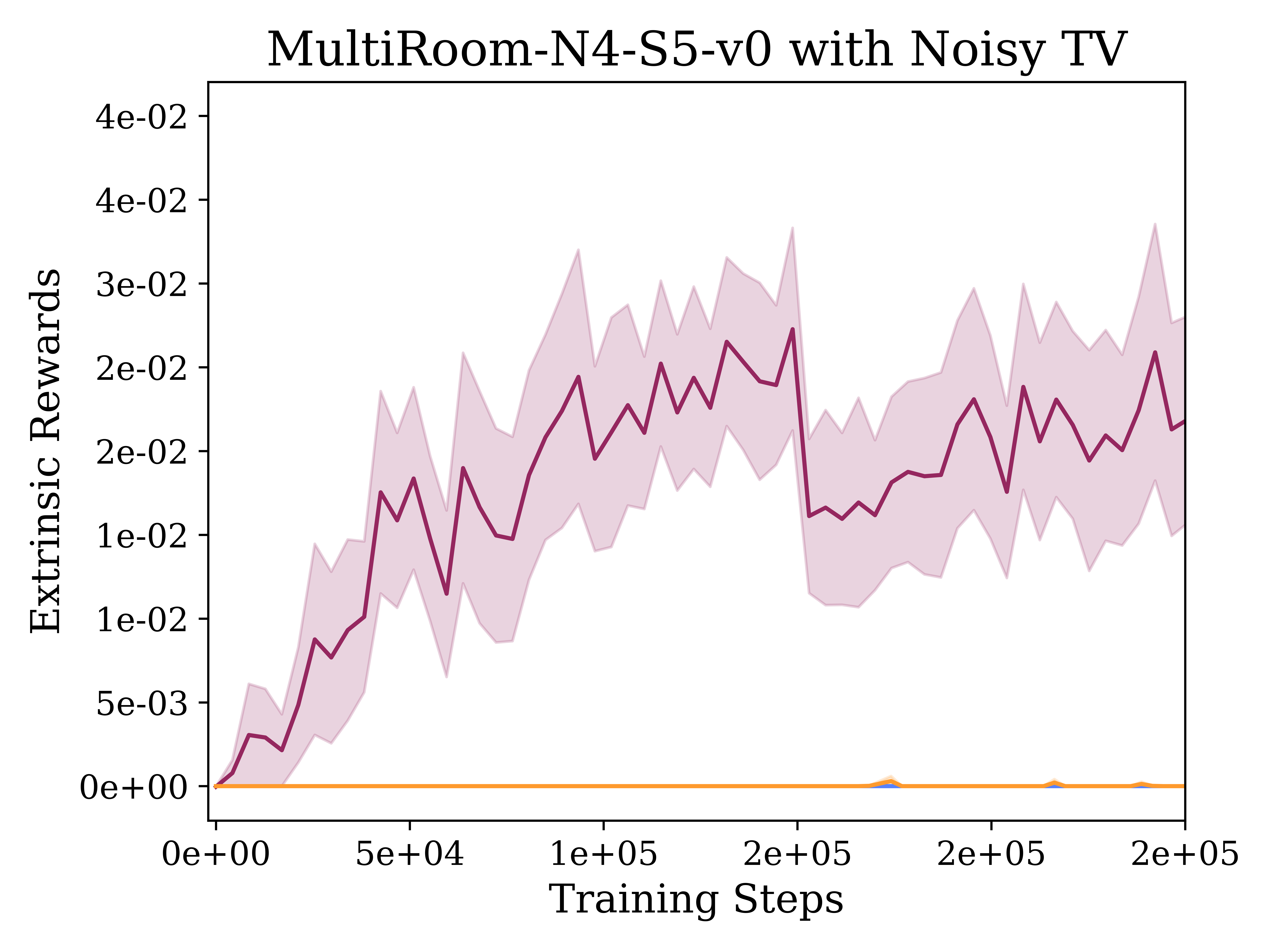}
\caption{} 
\end{subfigure}
\bigskip 
\begin{subfigure}{0.49\textwidth}
\includegraphics[width=\linewidth]{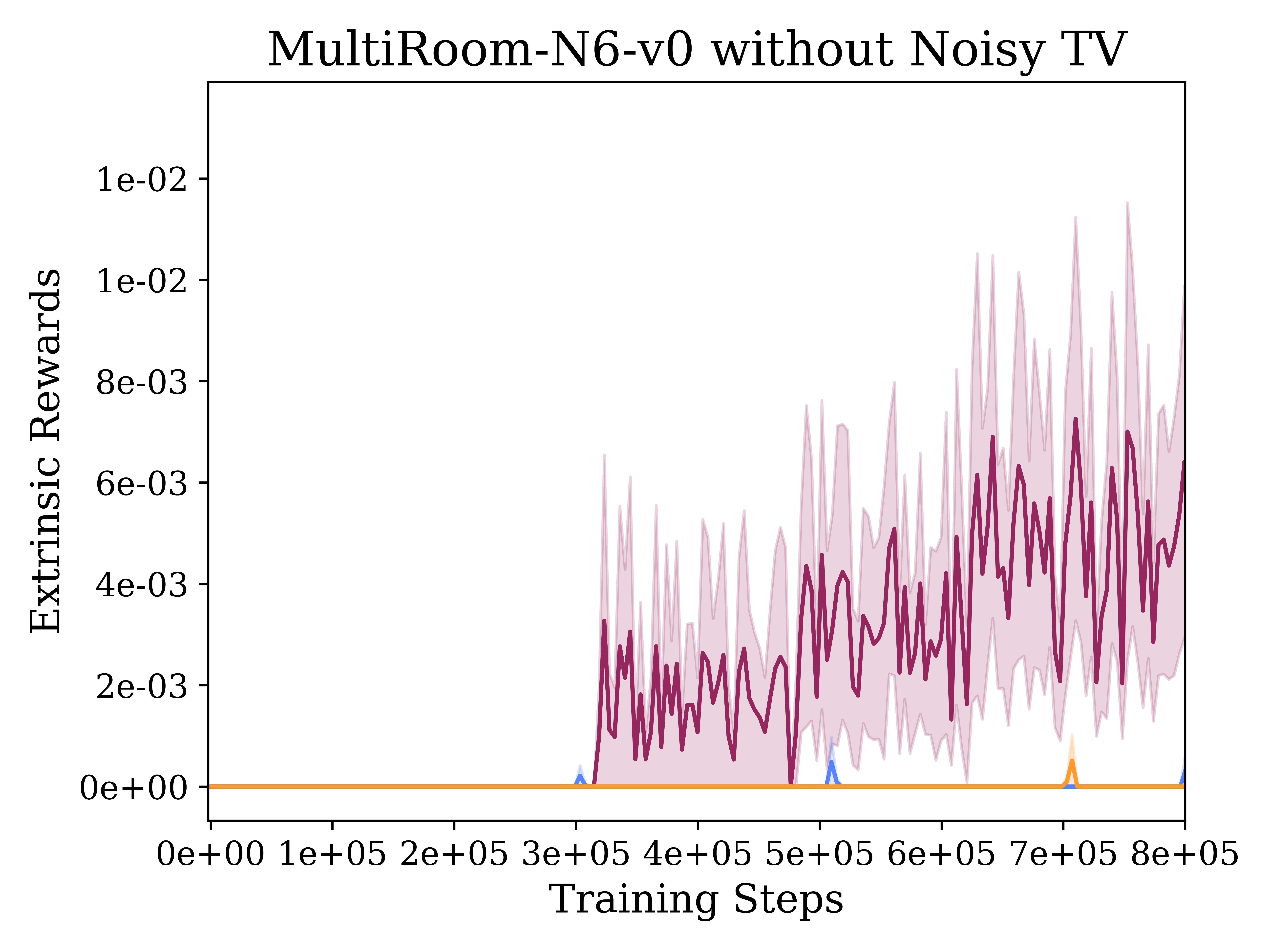}
\caption{} 
\end{subfigure}
\hfill
\begin{subfigure}{0.49\textwidth}
\includegraphics[width=\linewidth]{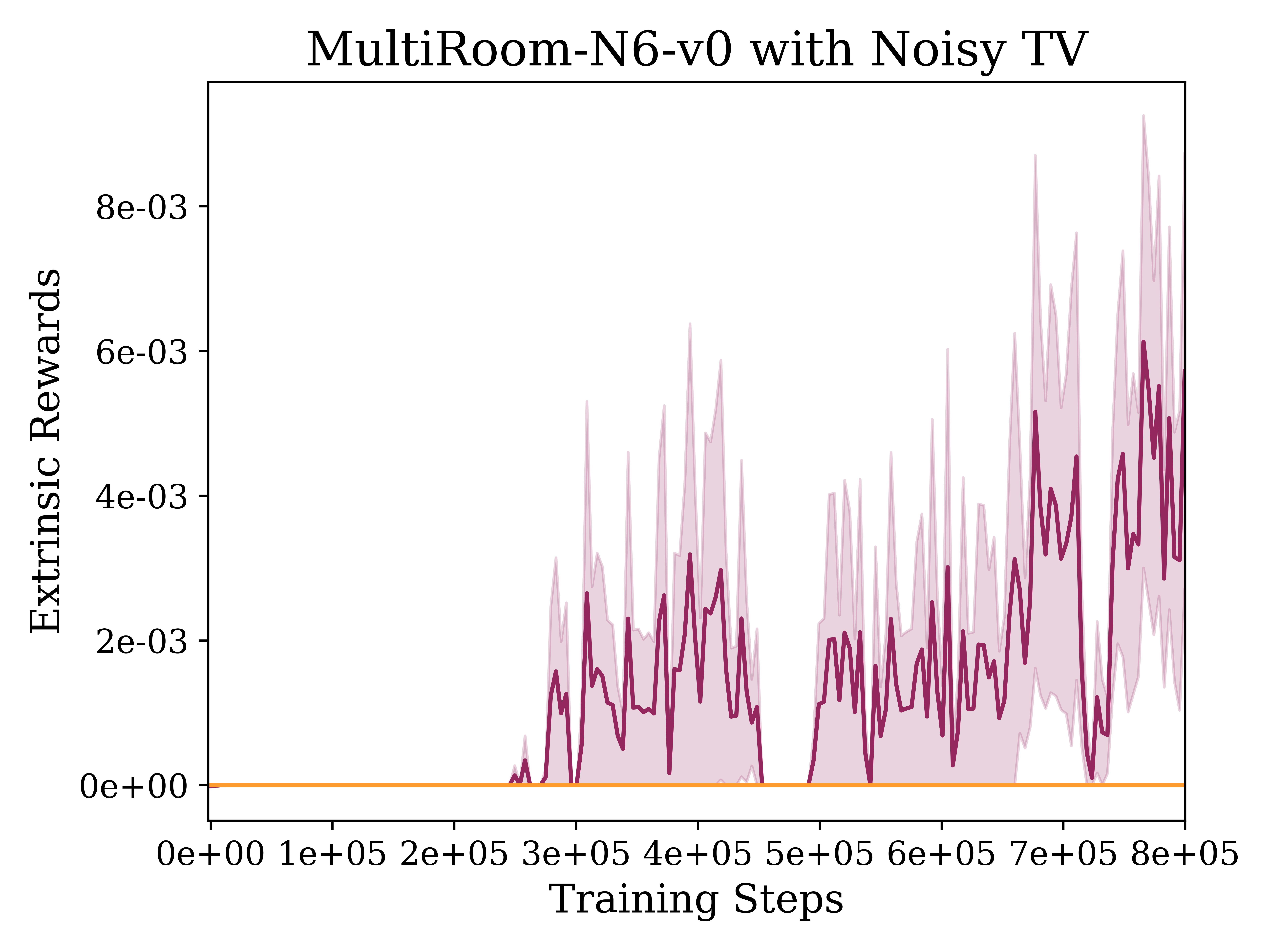}
\caption{} 
\end{subfigure}
\end{minipage}
\caption{AMA is the only approach that can achieve significant extrinsic rewards both with and without a noisy TV present in minigrid. Mean squared error intrinsic rewards are also able to achieve significant extrinsic rewards on the 4 room environment without the noisy TV present. Interestingly, aleatoric mapping agents are the only method that is able to receive significant extrinsic rewards on the 6 room environment even when no noisy TV is present, suggesting aleatoric mapping agents may improve exploration behaviours even when a prominent noisy TV is not present---this is consistent with the novel states visited results. One reason could be that the partial observability of the minigrid environments mean that one step predictions are sometimes unpredictable. For example, an empty observation that has no objects or walls may be presented to the agent in many different positions making the prediction of the next observation impossible.}
    \label{fig:minigrid_extrinsic_rewards}
\end{figure*}

Additional results showing the extrinsic rewards achieved by different agents in gym minigrid is shown in figure \ref{fig:minigrid_extrinsic_rewards}.

\begin{table}
\centering
\begin{tabular}{ |p{5cm}|p{5cm}|  }
 \hline
 Hyperparameter & Value\\
 \hline
 AMA learning rate & 0.001 \\
 MSE learning rate & 0.0001 \\
 Disagreement learning rate & 0.001 \\
 Random Network Distillation learning rate & 0.001 \\
 RMS Prop $\alpha$ & 0.99 \\
 RMS Prop $\epsilon$ & 1.000e-8 \\
 number of actors & 16 \\
 unroll length & 5 \\
 discount factor $\gamma$ & 0.99\\
 policy learning rate & 0.001\\
 GAE $\lambda$ & 0.95\\
 entropy coefficient & 0.01\\
 value loss coefficient & 0.5\\
 max grad norm & 0.5\\
 AMA uncertainty budget $\lambda$ & 0.1 \\
 AMA uncertainty coefficient $\eta$ & 1 \\
 \hline
\end{tabular}
\end{table}

\begin{figure*}
    \centering
    \includegraphics[width=0.5\linewidth]{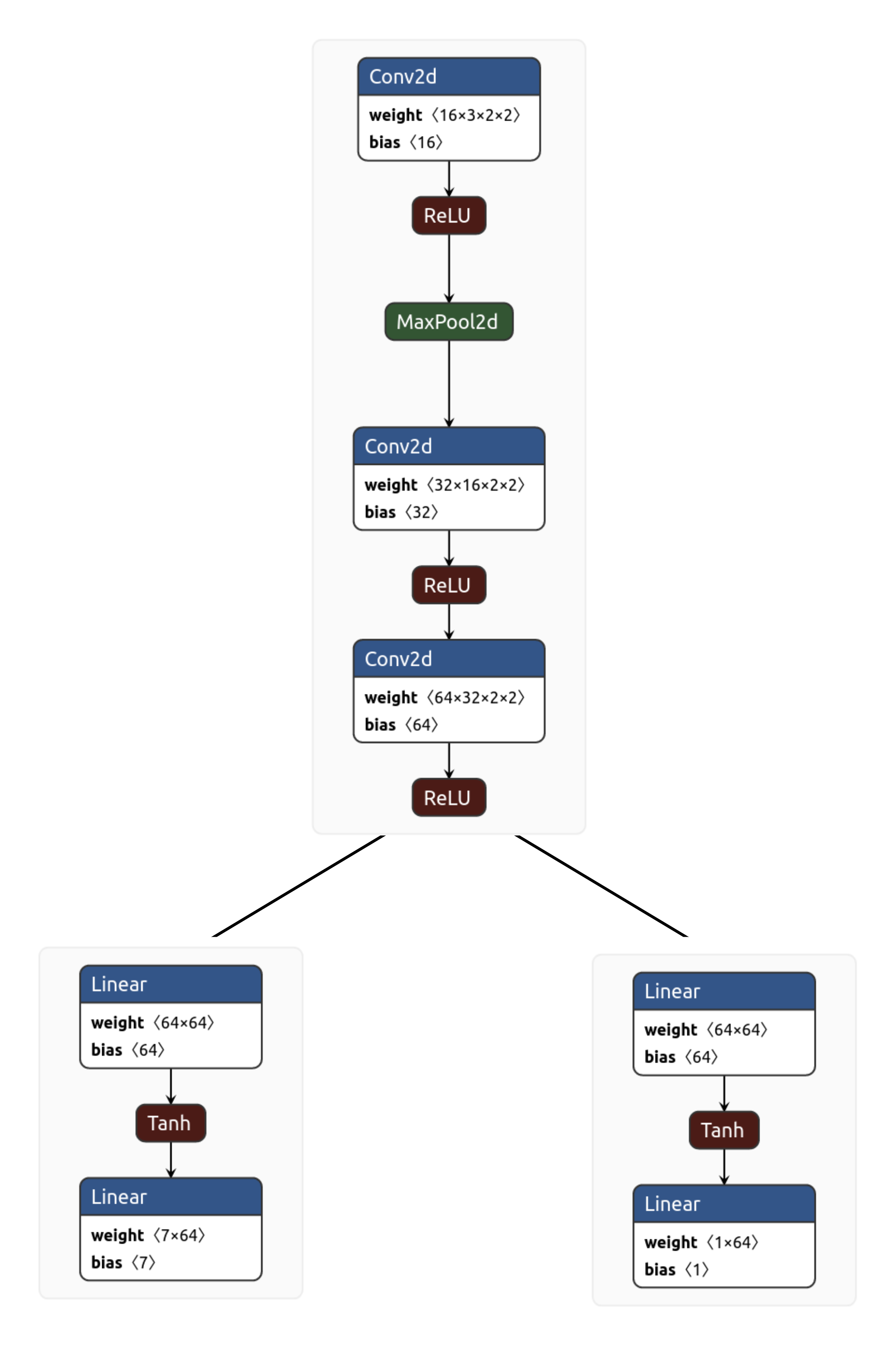}
    \caption{Actor critic architecture for the policy network in the minigrid experiments.}
\end{figure*}

\begin{figure*}
    \vspace{-3mm}
    \includegraphics[width=\linewidth]{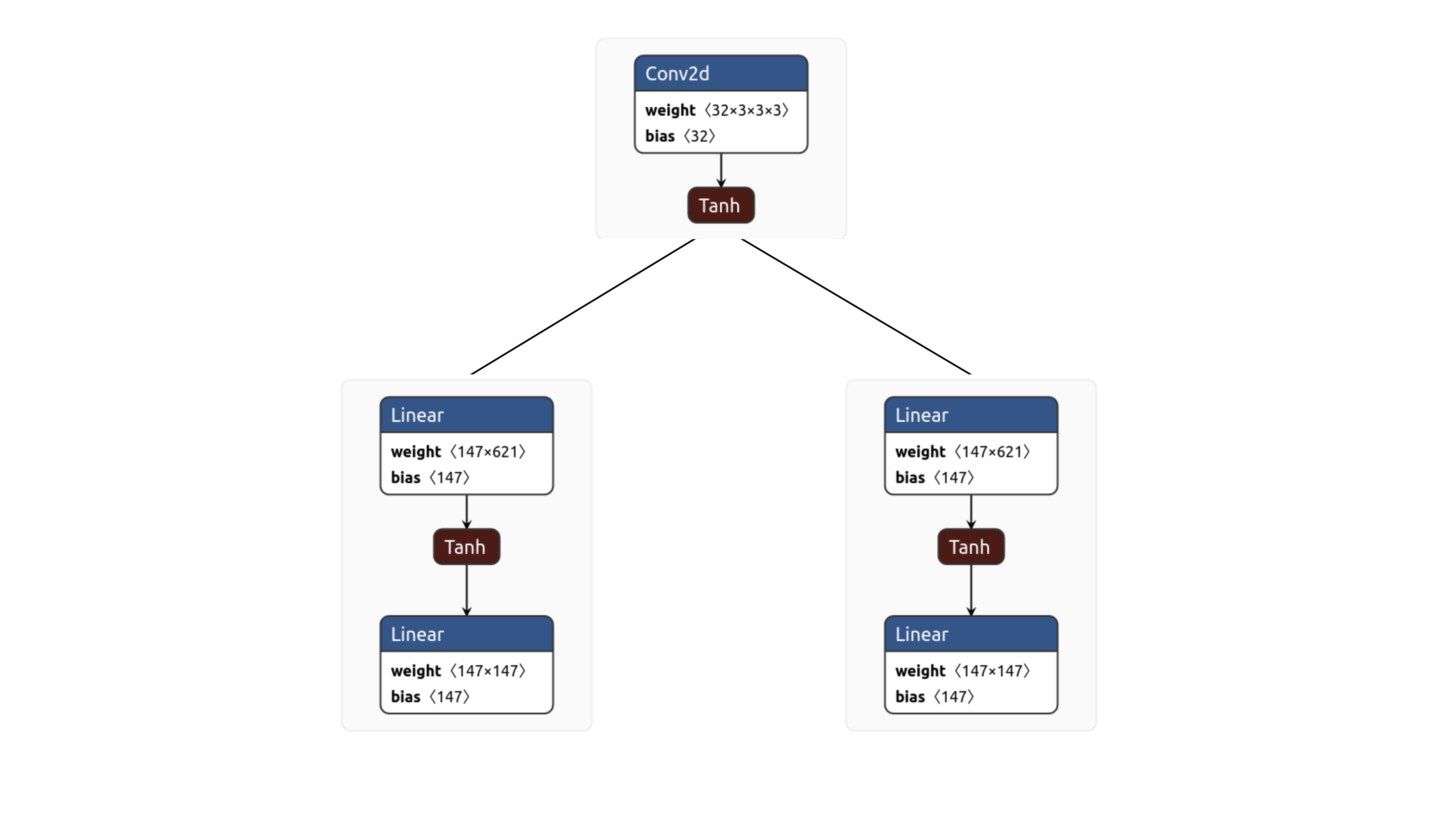}
    \caption{Curiosity forward prediction architecture for the minigrid experiments. For the MSE baseline the variance predictions are not used and loss is computed via a standard MSE.} 
\end{figure*}

\subsection{Atari}\label{retro_hyperparameters}
For each run we used five seeds. We use the official implementations \cite{burda2018exploration, burda2018large} for the baselines we compare to (with their default hyperparameters). For AMA and Pixel MSE we adapt from \cite{burda2018large}. The hyperparameters used for our AMA experiments can be found in Table 5. We did not change the PPO/vanilla curiosity hyperparameters from the original implementation we adapted and only changed the AMA hyperparameters. The hyperparameters were chosen by first exploring different configurations on smaller minigrid environments and evaluating promising configurations on the Atari environments.  

\begin{table}\label{ppo_hyperparams}
\centering
\begin{tabular}{ |p{6cm}|p{5cm}|  }
 \hline
 Hyperparameter & Value\\
 \hline
 global learning rate & 0.0001 \\
 normalise rewards & True \\
 number of PPO epochs & 3 \\
 number of actors & 128 \\
 unroll length & 128 \\
 discount factor $\gamma$ & 0.99\\
 GAE $\lambda$ & 0.95\\
 entropy coefficient & 0.001\\
 value loss coefficient & 0.5\\
 policy gradient clip range & [-1.1, 1.1] \\
 Pixel AMA uncertainty budget $\lambda$ & 1 \\
 Pixel AMA uncertainty coefficient $\eta$ & 2 \\
 IDF AMA uncertainty budget $\lambda$ & 1 \\
 IDF AMA uncertainty coefficient $\eta$ & 1 \\
 \hline
\end{tabular}\caption{Retro game policy hyperparameters}
\end{table}

The UNet architecture used for the forward predictions is described below. We duplicate the decoder head to create a two headed output but we only describe the encoder and decoder here. For the Pixel MSE baselines we use identical architectures but don't use the uncertainty predictions and train on MSE only. There are UNet style residual connections between the corresponding encoder and decoder layers. Leaky ReLU activations are used in the encoder layers and Tanh activations are used in the decoder layers. Batch normalisation is used throughout the hidden layers. Action information is concatenated at each layer. See supplementary code for further details.
\begin{table}
    \centering
    \begin{tabular}{ |p{3cm}|p{3cm}|p{3cm}|p{3cm}|  }
 \hline
 Layer Type & Filters & Kernel Size & Stride \\
 \hline
 Conv2d & 32 & 8 & 3 \\
 \hline
 Conv2d & 64 & 8 & 2 \\
 \hline
 Conv2d & 64 & 4 & 2 \\
 \hline
 Dense (512 Units) & N/A & N/A & N/A \\
 \hline
 Conv2d Transpose & 64 & 4 & 2 \\
 \hline
 Conv2d Transpose & 32 & 8 & 2 \\
 \hline
 Conv2d Transpose & 4 & 8 & 2 \\
 \hline
\end{tabular}\caption{Retro game forward prediction hyperparameters}
\end{table}

To integrate AMA into the IDF approach, we did not share any representations between mean and variance prediction heads, instead we used two prediction MLPs for the mean and variance. Leaky ReLU is used throughout hidden layers and action information is concatenated at each layer. We used five layers with 512 units each and UNet style residual connections. 

\subsection{Bandit}
We performed 3 repeats to produce the standard error regions show in the graph. Learning rate was tuned by hand, observing how well the network performed in making predictions as the bandit sampled different regions of the environment. The intrinsic reward method for the epistemic bandit is based on \cite{pathak19disagreement, lakshminarayanan2016simple}. We use an action value based bandit algorithm with $\epsilon$-greedy exploration \cite{sutton2018reinforcement}(p. 31).
\begin{table}
    \centering
    \begin{tabular}{ |p{5cm}|p{5cm}|  }
     \hline
     Hyperparameter & Value\\
     \hline
     epistemic learning rate & 0.0001 \\
     epistemic batch size & 32 \\
     aleatoric learning rate & 0.001 \\
     aleatoric batch size & 1000 \\
     aleatoric uncertainty budget $\lambda$ & 1 \\
     aleatoric uncertainty coefficient $\eta$ & 1 \\
     $\epsilon$ greedy $\epsilon$ & 0.1 \\
     \hline
    \end{tabular}\caption{Bandit hyperparameters.}
\end{table}

\begin{figure*}
    \vspace{-3mm}
    \includegraphics[width=\linewidth]{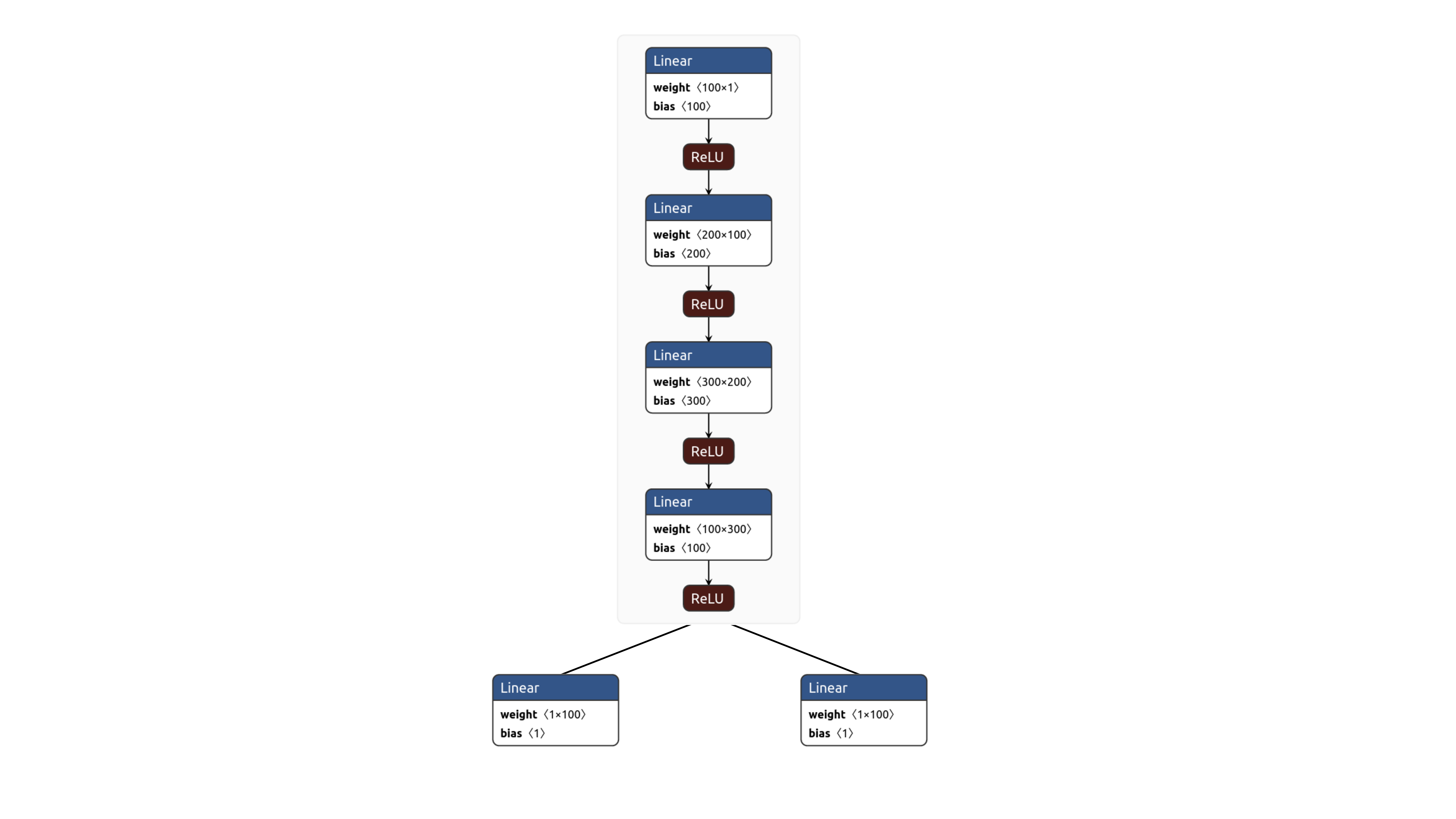}

    \caption{AMA prediction network for bandit tasks.} 
\end{figure*}

\begin{figure*}
    \vspace{-3mm}
    \begin{center}
    \includegraphics[width=0.25\linewidth]{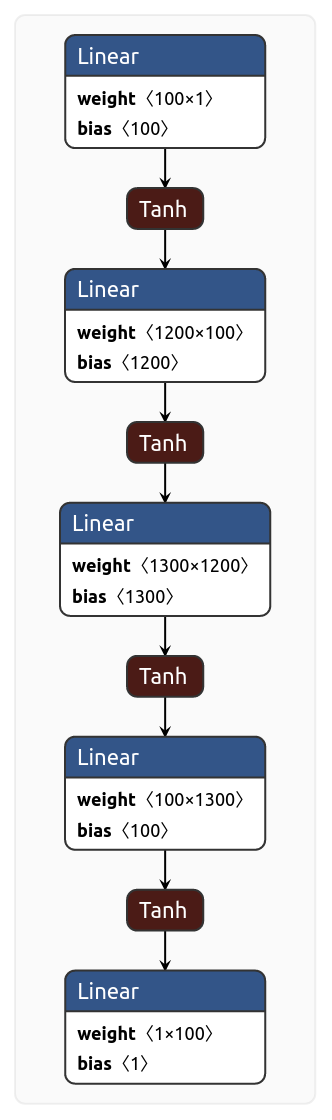}

    \caption{Epistemic prediction network for the bandit task.} 
    \end{center}
\end{figure*}

\subsection{Different Noise Distribution for Retro Games}\label{CIFAR}
For the CIFAR noisy TV we tiled a random CIFAR image (from the training set)
for each frame observed on the noisy TV. 
This required around around 2.5 tiles to fill the $84\times84$ pixels of the 
retro game frames. An example frame can be seen in Figure 12.

\begin{figure*}\label{CIFAR_noisy_TV}
    \vspace{-3mm}
    \begin{center}
    \includegraphics[clip, trim=3cm 11cm 3cm 0cm, width=0.5\textwidth]{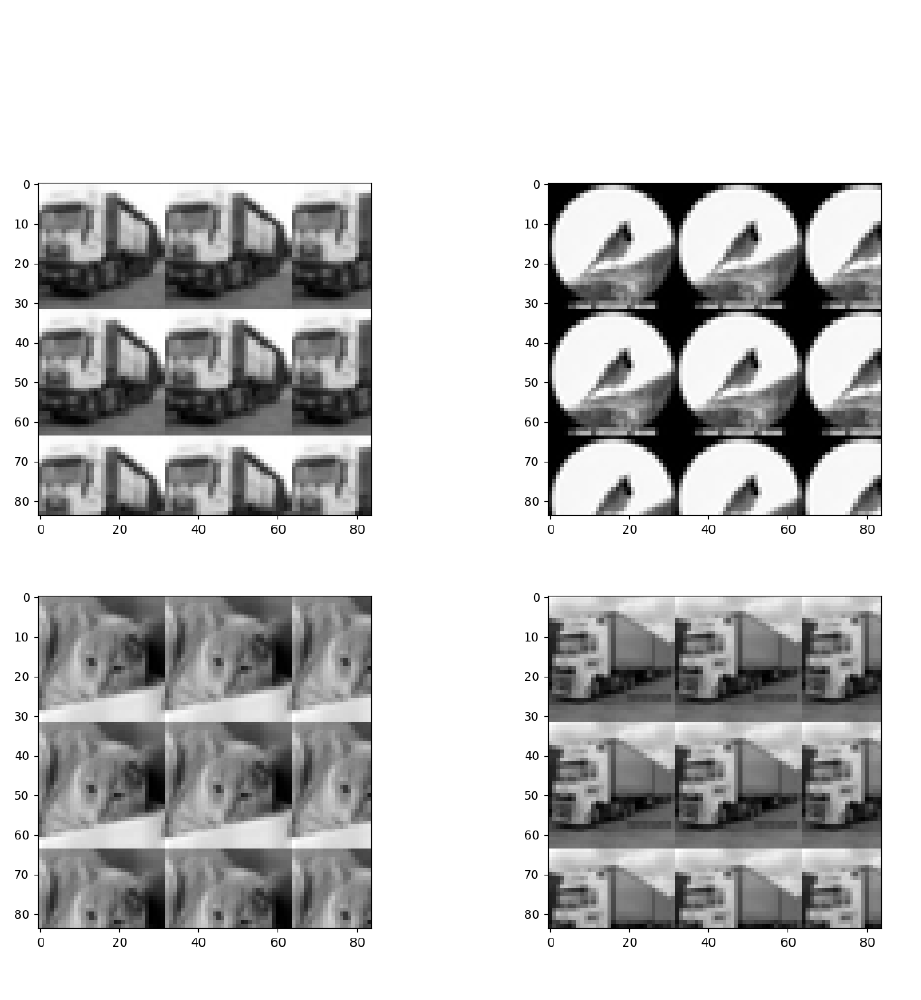}
    \caption{Four examples of frames that an agent might see when interacting with the 
    CIFAR noisy TV. This is a complex noise distribution picked to test the limits of the 
    heteroscedastic aleatoric uncertainty estimation.} 
    \end{center}
\end{figure*}

\subsection{Bank Heist Plotting}\label{bank_heist_x_explanation}
As briefly mentioned in the main text, the lines plotted for Bank Heist have approximate x points as the exact frames were not recorded directly with the average number of pixels covered in an episode. However, the logging step was recorded with the average pixel coverage. In the main text we use the frame count from the nearest recorded step to the step used for pixel coverage. To show the trends of these graphs show similar results we plot pixel coverage against training steps below. Note for all methods besides pixel AMA the Bank Heist plots are smoothed with following code snippet\footnote{\url{https://stackoverflow.com/a/11352216}} so that the crowded plot is readable.
\begin{figure}
\centering
\begin{subfigure}[b]{.32\linewidth}
\includegraphics[width=\linewidth]{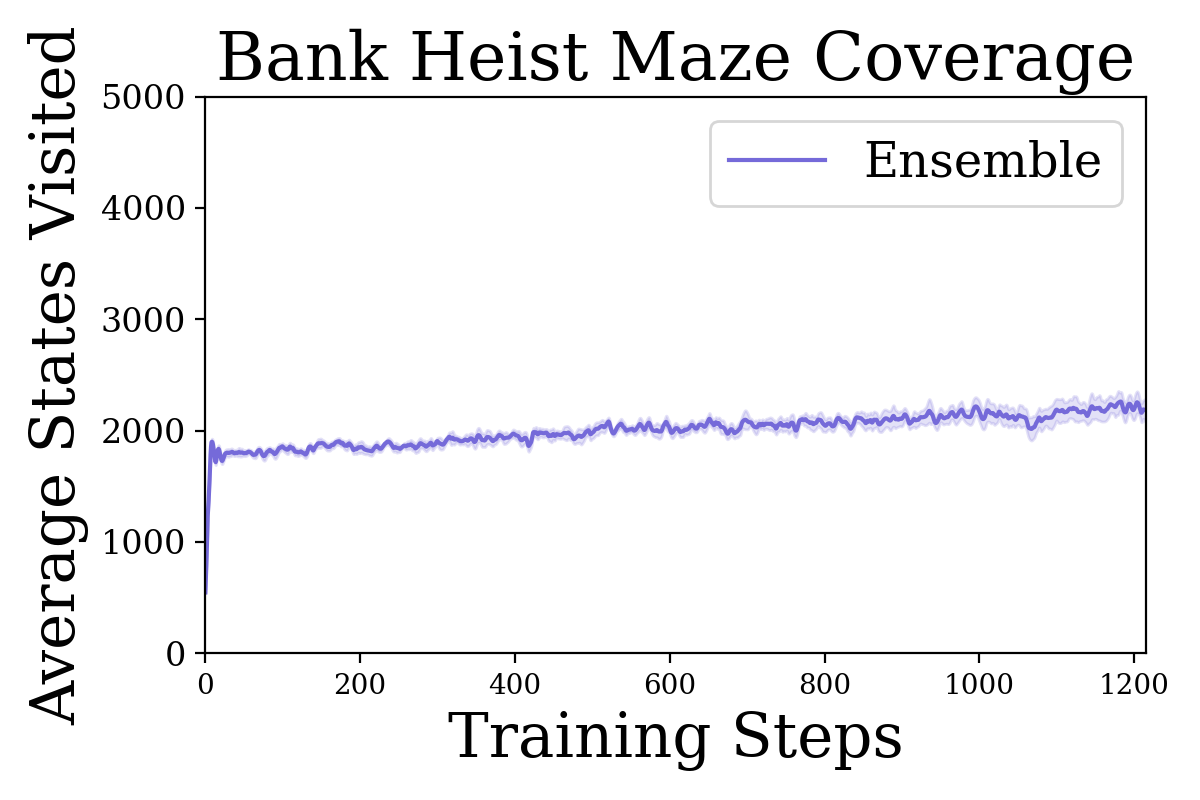}
\caption{}\label{fig:mouse}
\end{subfigure}
\begin{subfigure}[b]{.32\linewidth}
\includegraphics[width=\linewidth]{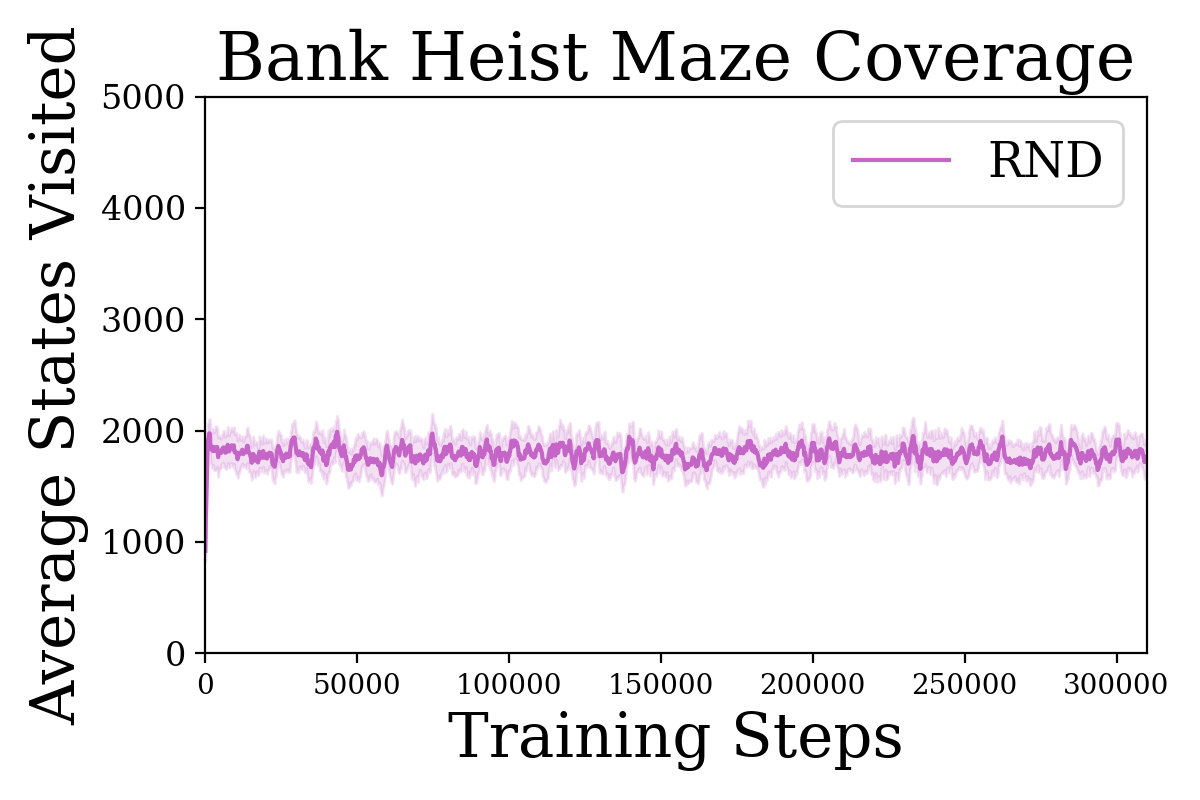}
\caption{}\label{fig:mouse}
\end{subfigure}
\begin{subfigure}[b]{.32\linewidth}
\includegraphics[width=\linewidth]{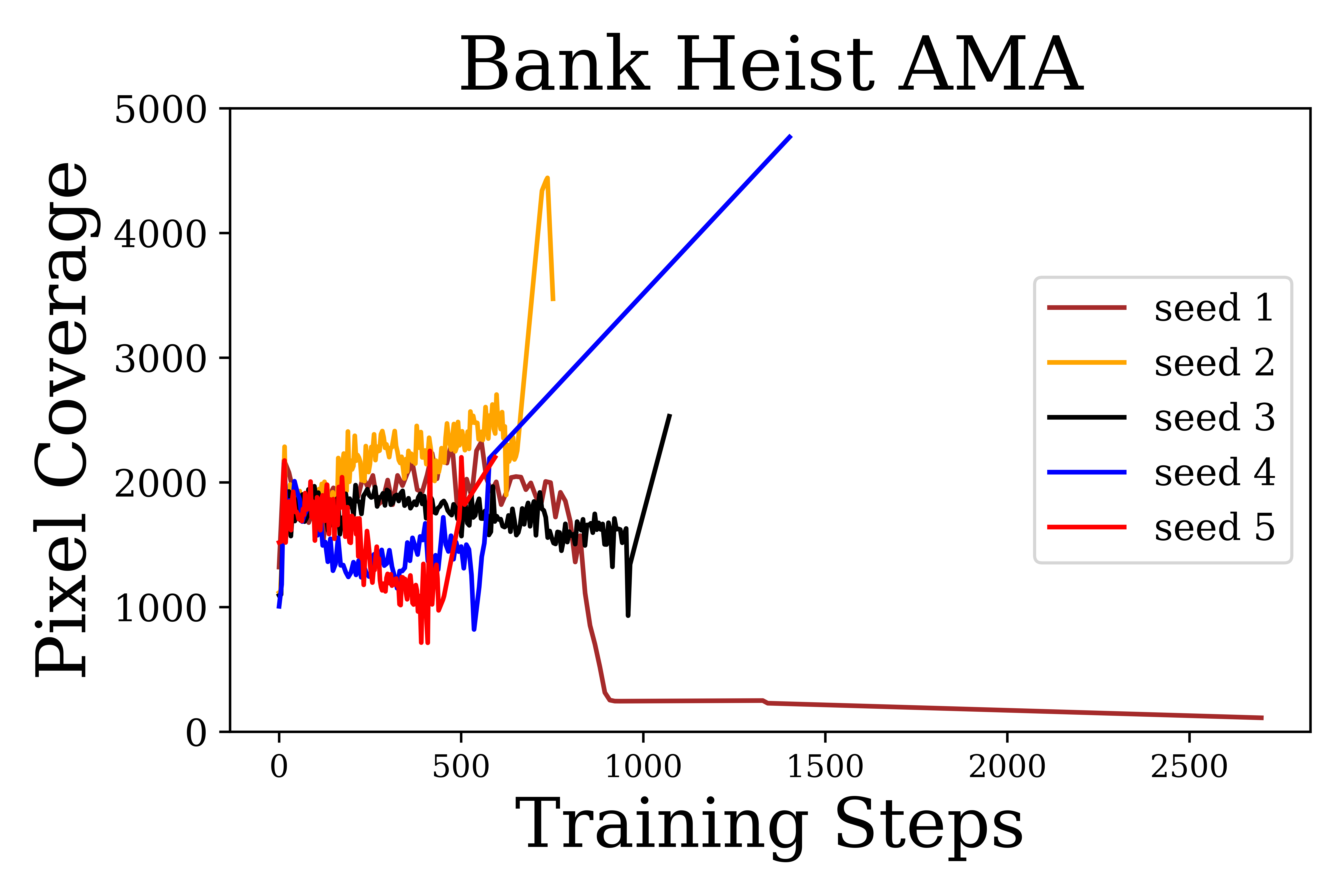}
\caption{}\label{fig:mouse}
\end{subfigure}

\begin{subfigure}[b]{.45\linewidth}
\includegraphics[width=\linewidth]{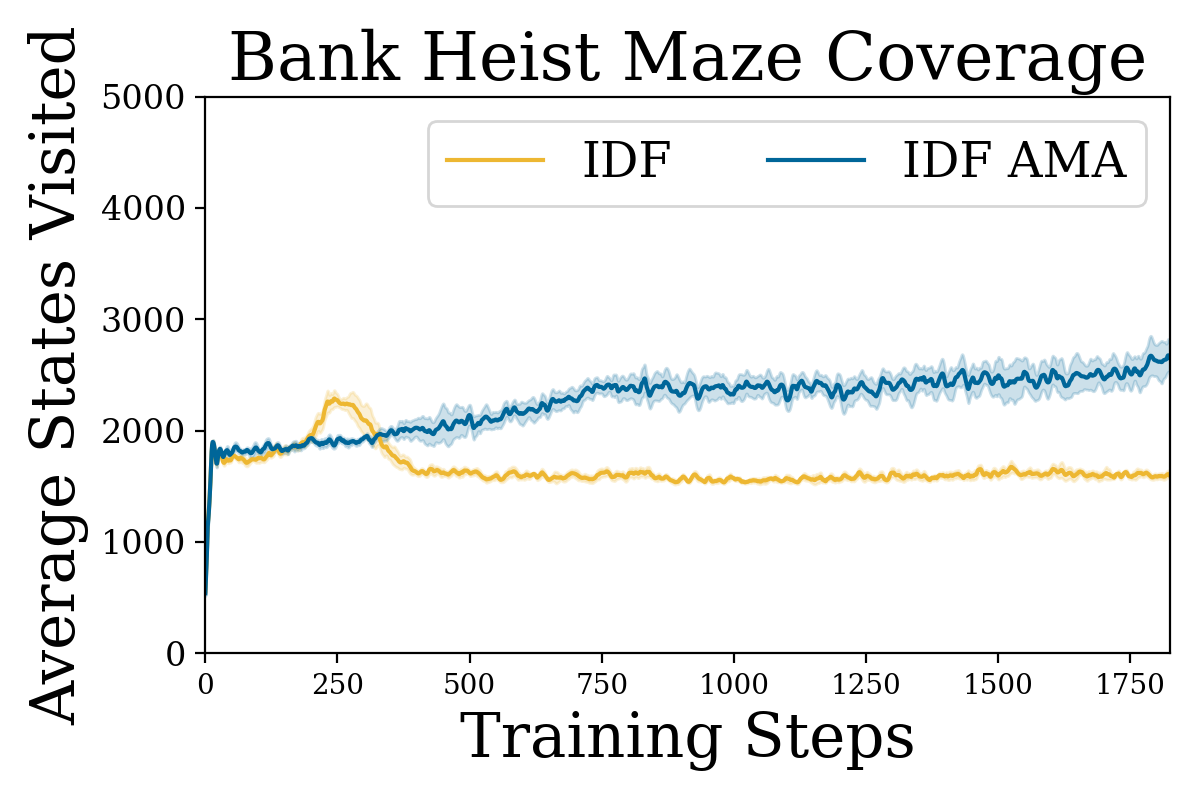}
\caption{}\label{fig:gull}
\end{subfigure}
\begin{subfigure}[b]{.45\linewidth}
\includegraphics[width=\linewidth]{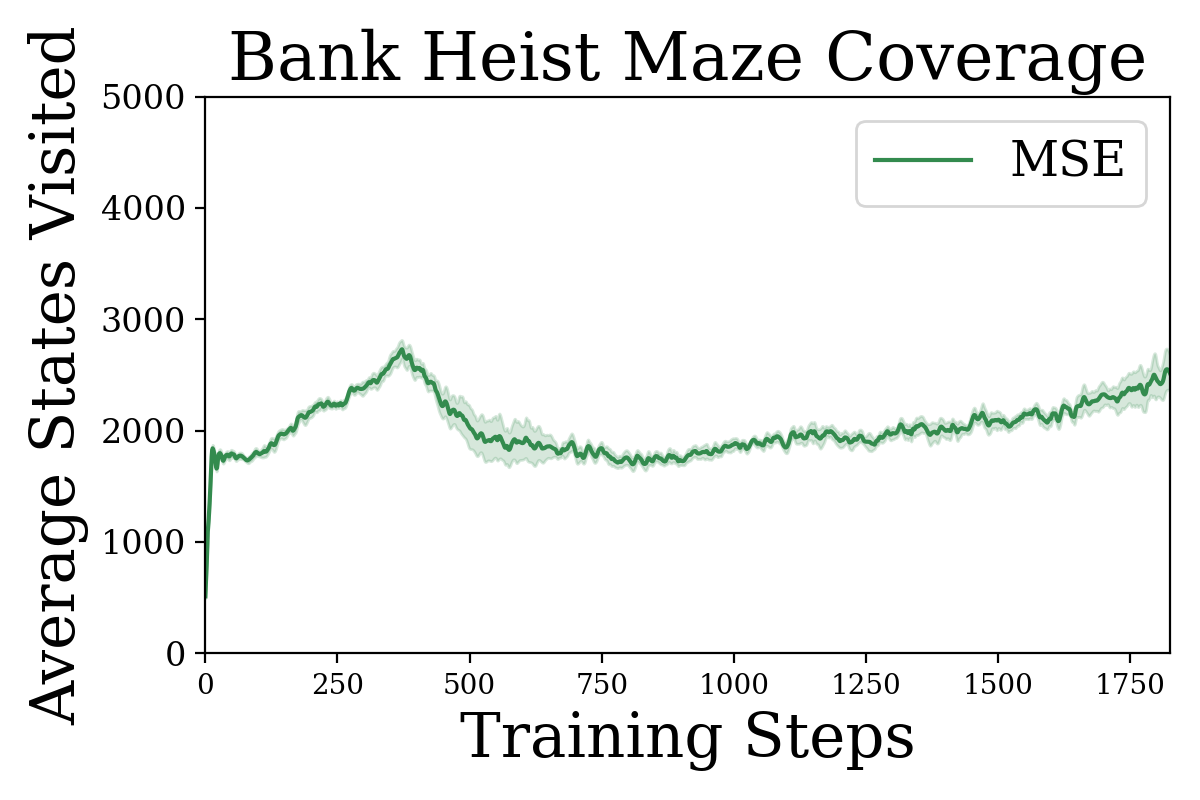}
\caption{}\label{fig:tiger}
\end{subfigure}
\caption{Trends of Bank Heist exploration plotted exactly against logging steps, showing similar trends to the approximate x-axis used in the main text. The training steps are not comparable between different methods. The pixel MSE results cannot be plotted together meaningfully against training steps as slightly different logging was used with different seeds.}
\label{fig:animals}
\end{figure}

\subsection{Eventual Decrease in Exploration During Mario Training}\label{mario}

In initial experiments we noticed the AMA agent lost motivation to explore its environment after reaching it peak extrinsic reward. To ensure that this 
was not an inherent problem with AMA, we ran the other curiosity 
methods for further frames and found similar eventual decreases in extrinsic reward
(Figure 14). Presumably the cause of this is that once a significant portion 
of the environment has been explored the agent is no longer motivated to 
return to those regions (as prediction error decreases)---this issue 
has been noted by previous authors \cite{pathak2017curiosity}.

\begin{figure*}
    \begin{subfigure}{0.48\textwidth}
        \includegraphics[width=\linewidth]{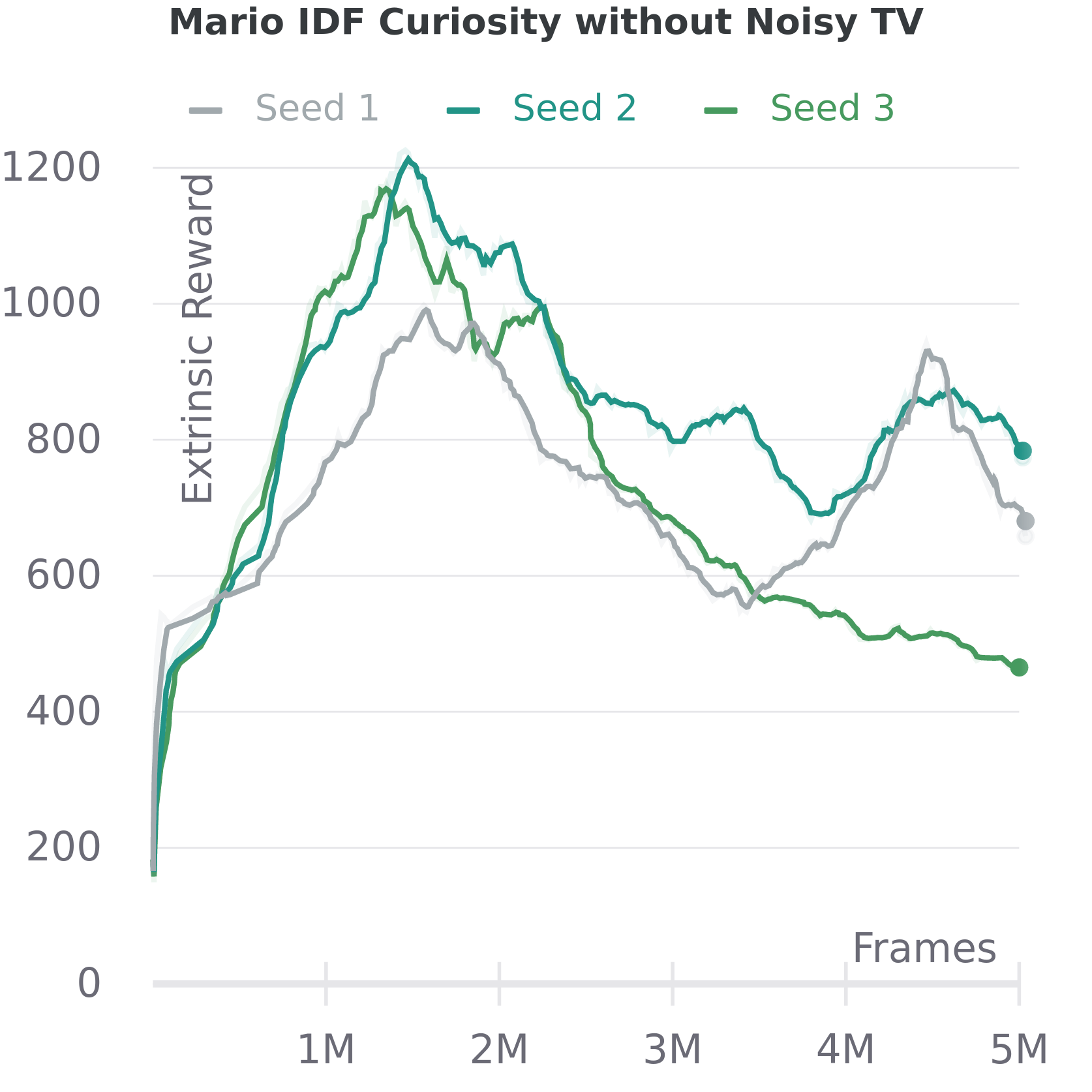}    
    \end{subfigure}
    \begin{subfigure}{0.48\textwidth}
        \includegraphics[width=\linewidth]{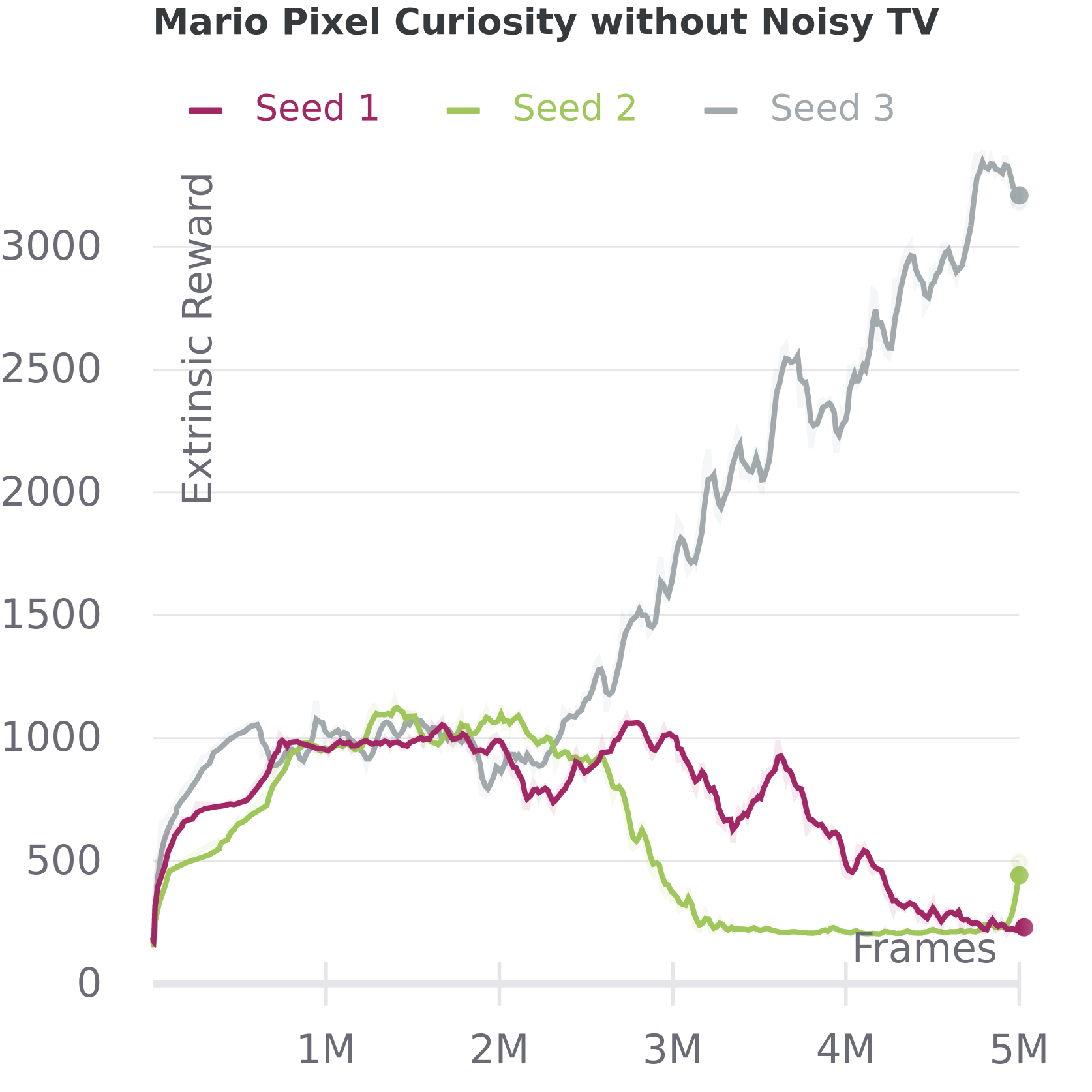}
    \end{subfigure}
    \caption{Repeats ran for more frames on Mario for MSE Pixel curiosity and MSE IDF curiosity. 3/3 of seeds see a decrease in performance for IDF curiosity and 2/3 in the pixel curiosity case.}

\end{figure*}\label{mario_decreasing}

\subsection{Potential Negative Social Impacts}
The work presented is very far from any real world deployment. If it were to be deployed in any real world context then extensive testing would need to be done to understand how the curiosity agents would behave in novel environments as erratic behaviours could be dangerous in, for example, a robotic control context. The AMA objective is overarching (like other curiosity methods) and so care should be taken if deploying in something like a recommender system as the agent could find certain behaviours intrinsically rewarding that you might not have intended it to (like the noisy TV problem). Lastly, although the AMA system contains notions of uncertainty quantification, that does not mean it is able to completely understand the limits of its predictions and so one should not be overconfident in its abilities to do so.

\subsection{Hardware}
The experiments were performed on three different machines depending on their availability: A 32 core CPU with one GeForce GTX TITAN X, a 12 core CPU with two GeForce GTX TITAN Xs and one 8 core CPU with two GeForce 2080Ti GPUs.

We list times here on the 2080Ti machine, the other machines were as much as $2\times$ slower. The minigrid experiments took around 40 minutes per run, the Space Invader experiments took around 12 hours per run, the Mario experiments took around 1 hour and 20 minutes per run.

\subsection{Licensing}
The repository from \cite{rl_starter_files} has an MIT license. The code used from \cite{gym_minigrid} has an Apache License 2.0. \cite{Raileanu2020RIDE:} has a creative commons license. Besides those listed we are not aware of any further code licensing. We adapted a copyright free rat silhouette image for the bandit figure \footnote{\url{https://pixabay.com/vectors/rat-rodent-silhouette-gold-chinese-5184465/}}. 

\subsection{Further Code Acknowledgements}
Although we did not directly use their code we would like to acknowledge the following open source contributions that provided a useful reference when implementing Kendall and Gal's \cite{kendall2017uncertainties} aleatoric uncertainty estimation algorithms:

\texttt{https://github.com/ShellingFord221/My-implementation-of-What-Uncertainties-\linebreak Do-We-Need-in-Bayesian-Deep-Learning-for-Computer-Vision}

\parbox{\linewidth}{\url{https://github.com/pmorerio/dl-uncertainty}}

\parbox{\linewidth}{\url{https://github.com/hmi88/what}}
\end{document}